\documentclass{article}
\usepackage[dblblindworkshop, final]{neurips_2025}

\usepackage[utf8]{inputenc} % allow utf-8 input
\usepackage[T1]{fontenc}    % use 8-bit T1 fonts
\usepackage{hyperref}       % hyperlinks
\usepackage{url}            % simple URL typesetting
\usepackage{booktabs}       % professional-quality tables
\usepackage{amsfonts}       % blackboard math symbols
\usepackage{nicefrac}       % compact symbols for 1/2, etc.
\usepackage{microtype}      % microtypography
\usepackage{xcolor}         % colors

\usepackage{amssymb}
\usepackage{amsmath}
\usepackage{graphicx}
\usepackage{subfigure}
\usepackage{listings}
\usepackage[most]{tcolorbox}
\tcbset{
  colback=blue!5!white,
  colframe=blue!70!black,
  boxrule=0.8pt,
  arc=3pt,
  left=6pt,
  right=6pt,
  top=4pt,
  bottom=4pt,
  enhanced,
  breakable
}

\usepackage{kotex}

\title{A-LAMP: Agentic LLM-Based Framework for Automated MDP Modeling and Policy Generation}
\workshoptitle{Multi-Turn Interactions in Large Language Models}

\author{%
  Hong Je-Gal\\
  Department of AI \& Robotics\\
  Sejong University\\
  Seoul 05006 \\
  \texttt{jagrhong@sju.ac.kr} \\
  \And
  Chan-Bin Yi\\
  Department of AI\\
  Sejong University\\
  Seoul 05006\ \\
  \texttt{22012165@sju.ac.kr} \\
  \And
  Hyun-Suk Lee\thanks{Corresponding author.} \\
  Department of AI \& Robotics\\
  Sejong University\\
  Seoul 05006\ \\
  \texttt{hyunsuk@sejong.ac.kr} \\
}

\begin{document}

\maketitle

\begin{abstract}
Applying reinforcement learning (RL) to real-world tasks requires converting informal descriptions into a formal Markov decision process (MDP), implementing an executable environment, and training a policy agent. Automating this process is challenging due to modeling errors, fragile code, and misaligned objectives, which often impede policy training. We introduce an agentic large language model (LLM)-based framework for automated MDP modeling and policy generation (A-LAMP), that automatically translates free-form natural language task descriptions into an MDP formulation and trained policy. The framework decomposes modeling, coding, and training into verifiable stages, ensuring semantic alignment throughout the pipeline.
Across both classic control and custom RL domains, A-LAMP consistently achieves higher policy generation capability than a single state-of-the-art LLM model. Notably, even its lightweight variant, which is built on smaller language models, approaches the performance of much larger models. Failure analysis reveals why these improvements occur. In addition, a case study also demonstrates that A-LAMP generates environments and policies that preserve the task's optimality, confirming its correctness and reliability.

% These results highlight that structured decomposition--rather than sheer model scale--is the key enabler of reliable and cost-efficient policy automation.

\end{abstract}

\section{Introduction}
Reinforcement learning (RL) has emerged as a foundational paradigm for sequential decision making in dynamic and uncertain environments ~\citep{kaelbling1996reinforcement,bertsekas2008neuro}. 
It enables agents to acquire  
%
% optimal behaviors in each state 
%
a good policy to achieve a given goal
through trial-and-error interactions with their environments, guided by appropriate reward signals  ~\citep {watkins1992q,sutton1999policy, konda1999actor}.
Modern deep RL algorithms have shown impressive performance in various domains such as robotics, autonomous control, and strategic gameplay \citep{vinyals2019grandmaster, silver2018general, berner2019dota}.

%--
To make RL effective in real-world conditions, it is essential to establish a well-posed Markov decision process (MDP) that specifies what to optimize and how actions influence outcomes as a formal bridge between informal task descriptions and an executable environment with a training loop that generates a deployable policy. 
%Decision-making tasks aimed at achieving specified goals should therefore be translated into mathematical formulations and executable software.
%: a well-posed MDP that specifies what to optimize and how actions influence outcomes and an executable environment with a training loop that generates a deployable policy. 
In practice, however, transitioning from such an MDP specification to a capable policy requires maintaining intact the semantics throughout training the policy using the environment: the information encoded in the state must appear in the observations, the implemented choices of the agent must match the allowable actions, the environment must realize the assumed dynamics and reward, and the training loop must optimize precisely that objective under the given constraints. 
Only when this end-to-end alignment is preserved and made verifiable through evaluation and traceable logs, the resulting policy can reliably reflect the target task. However, in real-world deployments for various tasks, this alignment is fragile and may pose persistent challenges.

One of the major challenges is deriving a precise MDP from real-world tasks. To this end, policy designers must decide how to abstract the target task into a mathematical form by defining informative states, identifying feasible actions, and translating the task's objectives into reward signals that actually drive policy learning. 
Another challenge is implementing an environment in which to train a policy that reflects the MDP formulation of the task.
This process becomes more difficult when these elements are buried in unstructured or domain-specific artifacts, such as simulator configuration files, engineering documentation, or informal natural language specifications. This makes the process manual and expertise-intensive. 
As a result, constructing an MDP and creating an executable environment are often time-consuming, error-prone, and dependent on human experts, creating a barrier to RL adoption in domains such as network resource scheduling, industrial automation, and supply chain optimization.~\citep{ye2019deep, luong2019applications}.

Furthermore, this traditional RL process conducted by human experts lacks flexibility when tasks or parameters change since learned policies often cannot be reused directly ~\citep{taylor2009transfer,finn2017model, rakelly2019efficient, rusu2016progressive}.
Even minor modifications, such as switching the task objective from maximizing performance to minimizing energy consumption in the same wireless network, require experts to reformulate the MDP manually and retrain a new policy from scratch.
This inflexibility severely limits the scalability and efficiency of RL workflows because even experienced experts must repeatedly translate vague problem semantics into coherent MDP components and reimplement similar structures across tasks with only minor variations.

To address these challenges, we introduce an agentic large language model (LLM)-based framework for automated MDP modeling and policy generation (A-LAMP). A-LAMP is a modular multi-agent LLM framework that formally establishes an MDP from a free-form natural language description. It also generates policy training code, including an executable RL environment, which can be used to train policies.
A-LAMP reliably processes this end-to-end automation from the natural language description to policy generation while preserving interpretability for human experts by orchestrating specialized agents.
%
%This is possible because A-LAMP explicitly modularizes each step of the RL pipeline, producing either a human-readable description or an equation-level formulation.
%
%Since the entire process--from the natural language description to policy generation--is fully automated end-to-end, adapting to new objectives or simulator changes only requires an updated description, not extensive re-engineering.
As a result, it lowers the barrier to deploying RL in real-world settings. The key contributions of A-LAMP are summarized below:
\begin{itemize}
    \item \textbf{Enhanced capability in MDP modeling and policy generation}: A-LAMP decomposes the MDP formulation and policy generation process into specialized LLM agents, constructing precise mathematical representations and corresponding policy structures. This approach yields more reliable and consistent policies than a single large LLM (e.g., GPT-4o).
    \item \textbf{Transparency and interpretability}: Each component of the MDP--such as inferred objectives, decision variables, and constraints--is modularized in A-LAMP, producing either a human-readable description or an equation-level formulation. This design allows experts to inspect, validate, and refine the modeling process at any stage.
    \item \textbf{Adaptability across tasks and environments}: A-LAMP automatically generates both MDPs and policies from free-form natural language descriptions. This enables new tasks or changes to existing tasks to be handled through automated regeneration based on updated descriptions, rather than complete manual re-engineering.
    \item \textbf{Enhancing productivity via automation}: A-LAMP significantly improves productivity and lowers the expertise barrier for RL deployment by automating the labor-intensive steps such as environment construction, reward specification, and policy initialization.
\end{itemize}

\section{Related Work}

Our proposed framework, A-LAMP, leverages three core capabilities of LLMs: (1) general reasoning and agentic decision-making, which exploits the LLMs' ability to perform structured reasoning, high-level planning, and multi-agent orchestration, (2) policy and reward design for RL, which is based on the LLMs' capacity to interpret task goals and translate them into well-defined objectives, and reward functions, (3) understanding of the environment and generation of code, which requires the LLMs' ability to transform formalized task specifications into executable RL environments. The following subsections review related previous work and situate A-LAMP within the broader research landscape.\looseness=-1

% \subsection{LLMs for General Reasoning and Agentic Decision-Making}
\noindent\textbf{General reasoning and agentic decision-making:}
LLMs have demonstrated strong capabilities in general reasoning, symbolic manipulation, and structured decision-making across various domains.
Foundational studies such as Sparks of AGI~\citep{bubeck2023sparks} and the GPT-4 Technical Report~\citep{openai2023gpt4} illustrate that LLMs can perform high-level planning and control-oriented reasoning.
Generative agents~\citep{park2023generative} show that LLMs can simulate believable multi-agent behaviors.
To extend these abilities, agentic prompting methods, e.g., Toolformer~\citep{schick2023toolformer}, ChatDev~\citep{qian2023chatdev}, OpenAGI~\citep{pei2023dynamic}, and OptiMUS~\citep{ahmaditeshnizi2023optimus,ahmaditeshnizi2024optimus}--decompose workflows into specialized agents, each responsible for modular reasoning, planning, or verification.

% \subsection{LLMs for Reinforcement Learning: Policy and Reward Design}
\noindent\textbf{Policy and reward design for RL:}
Building on these reasoning capabilities, recent research explores how LLMs can help specific components of the RL pipeline.
EUREKA~\citep{ma2023eureka} shows that LLMs can autonomously design reward functions from natural language prompts, often outperforming human-crafted rewards.
SPRING~\citep{wu2023spring} transforms environment documentation into symbolic action plans, while PAL~\citep{gao2023pal} and Voyagers~\citep{wang2023voyager} employ LLM agents for interactive and logical planning.
The recent framework for (top-down) strategic planning~\citep{luytenstrategic} introduces a strategist agent that uses LLM-based strategy trees and reward shaping to improve the efficiency of exploration.
These studies highlight the growing trend of using LLMs as decision-making modules or reward designers in RL, while A-LAMP extends this direction by formalizing MDP components through agentic LLMs and generating capable policies.

% \subsection{LLMs for Environment Understanding and Code Generation}
\noindent\textbf{Environment understanding and code generation:}
Complementing their reasoning abilities, LLMs have excelled in code generation, a crucial skill for translating formalized task structures into RL environments.
Codex~\citep{chen2021evaluating} and CodeT5+~\citep{wang2023codet5+} have achieved strong performance in translating natural language to code.
More recently, G-Sim~\citep{holt2025g} combines LLM-driven structural reasoning with gradient-free calibration to build robust, causally grounded simulators.
Inspired by such approaches, A-LAMP organizes environment generation into a multi-agent LLM pipeline, extracting states, actions, and rewards and producing executable RL training code, thus bridging the gap between abstract task specifications and real-world policy learning.\looseness=-1

\section{A-LAMP: Agentic LLM-based Framework for Automated MDP Modeling and Policy Generation}

\subsection{Solving a Decision-Making Task via Policy Generation: A Human-Centered Perspective}
\label{sec: human_MDP}

From a human expert's perspective, generating a policy to solve a given task is not a linear technical process. Rather, it is an iterative process of reasoning that evolves from a conceptual understanding of the task to a mathematical formulation, and finally, to an executable implementation. This involves realizing the environment and reward specified by the MDP and optimizing the policy via the RL training loop.
When an expert encounters a new task, the initial challenge is to interpret the narrative description and to translate its linguistic cues into a coherent understanding of the situation, which involves identifying meaningful factors and parameters for decision-making. 
Throughout this process, the expert continually refines their evolving interpretation, drawing on three complementary sources: domain engineering knowledge, RL references, and the initial task idea.
%
% For example, in a wireless scheduling scenario, terms such as transmission power, bandwidth, or time slots are identified as critical components influencing system dynamics.

First, in an abstract idea phase, the expert assembles a conceptual structure of the task from the narrative of the task idea.
This clarifies the objective of the task--what success must mean in the given context (e.g., maximizing service quality with a fixed budget)--identifies the decision variables--the controllable options actually available to the agent (e.g., which resource to allocate at each decision step)--and surfaces the constraints--the non-negotiable limits that bound behavior (e.g., safety, capacity, or policy caps that must not be violated). 
These elements are proposed, cross-checked, and refined together. The integrated result forms the abstraction of the task idea.\looseness=-1
%
% For instance, determining which user transmits in each time slot might be a primary decision variable. At the same time, the practitioner accounts for physical, logical, or operational constraints that limit feasible actions, such as ``only one user can transmit at a time'' or ``the total transmission power must remain below a threshold.'' 
%
% This initial reasoning, progressing from problem comprehension to defining objectives, decision variables, and constraints, forms the conceptual foundation of the MDP.

After developing this conceptual structure, the human expert proceeds to a formulation phase. This phase involves encoding the problem as an MDP. 
During this phase, vague goals, such as ``maximize service quality,'' are translated into numerical objective functions, and constraints are expressed as algebraic inequalities or logical conditions. 
The state and action spaces are carefully defined to include only relevant and observable variables. 
In particular, finding the right balance is essential to this step. If the state is too complex, learning becomes difficult; conversely, if the state is too simple, important dynamics may be lost.
The formulation phase is often iterative and requires adjustments to achieve both fidelity to real-world dynamics while ensuring compatibility with RL algorithms.

\begin{figure}[!t]
    \centering
    \includegraphics[width=1.0\linewidth]{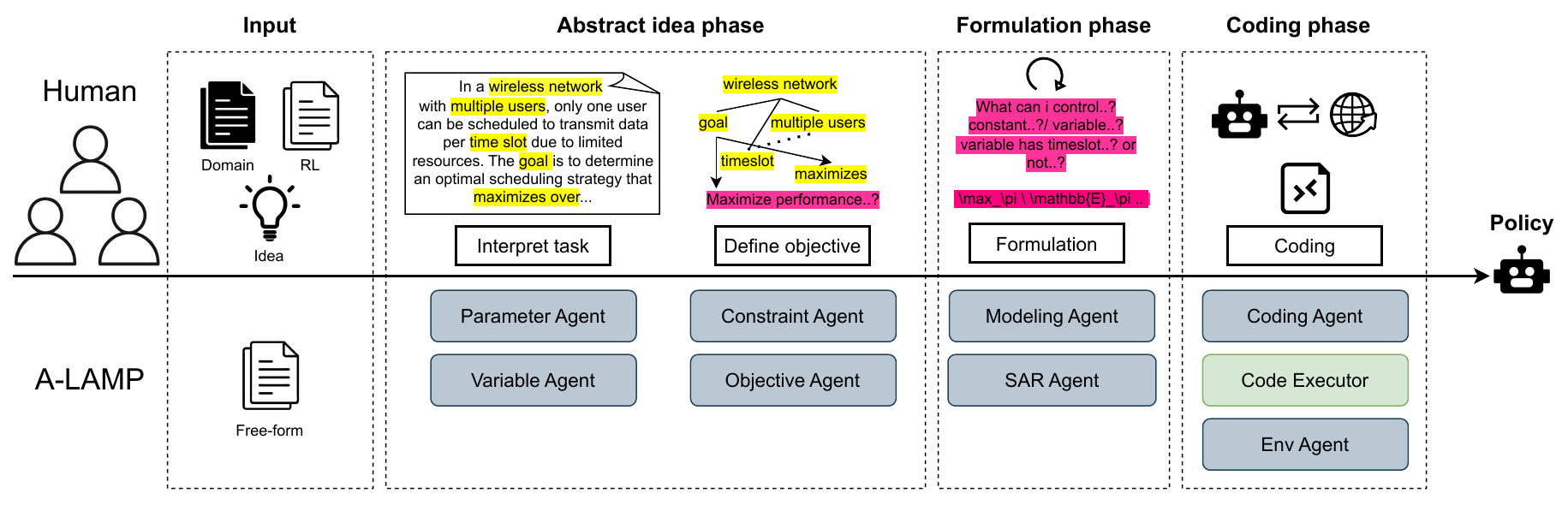}
    \caption{Comparison of policy generation processes: a manual human-expert pipeline (top) and the automated A-LAMP pipeline (bottom). Both follow three phases--abstract idea, formulation, and coding--to produce a policy. 
    Human experts consume domain knowledge, RL information/specifications, and a task idea as an input; A-LAMP takes a free-form natural-language description of the task as an input. A-LAMP replaces each cognitive step with specialized LLM agents.}
    \label{fig:human}
    \vskip -0.1in
\end{figure}

Finally, a coding phase bridges the gap between formal definitions and executable RL code. 
The human expert implements the MDP as code, mapping states to computable features, encoding the action set, and implementing reward logic that aligns with the intended optimization goal. 
This typically involves creating custom environments and integrating standard RL algorithms such as DQN or PPO. 
Since small implementation errors can lead to unstable training or misaligned behavior, the coding stage often includes a debugging loop. During this loop, the expert refines state definitions, reward functions, and constraints. 
The entire manual human expert pipeline for policy generation is depicted in Figure~\ref{fig:human}, from abstracting ideas to coding. This process is time-intensive and requires domain knowledge, mathematical reasoning, and programming proficiency.

\subsection{Automated MDP Modeling and Policy Generation via A-LAMP}
\label{sec:method}

A-LAMP automates three phases--abstract ideas, formulation, and coding--in the human policy generation process with specialized LLM agents, as illustrated in Figure \ref{fig:human}.
%
%We first systematize the human-expert RL process in Section \ref{sec: human_MDP} into a forward pipeline for controllability and efficiency across . 
%
In A-LAMP, each phase is further decomposed to ensure robustness across different model capabilities.
Such a narrower role of each agent lowers its reasoning burden and makes outputs easier to validate.
This enables even smaller LLMs to meet a baseline, while larger ones produce more precise results. 
The decomposed steps are ordered by their dependencies so that an agent runs only after the minimal prerequisites it needs have been extracted (e.g., parameters $\rightarrow$ objectives $\rightarrow$ decision variables $\rightarrow$ constraints $\rightarrow$ modeling $\rightarrow$ coding). 
%
%To ensure robustness across different model capabilities, we over-decompose tasks.
%
Finally, to preserve interpretability, each agent emits a transparent, human-readable outputs (schemas/equations/code), enabling stepwise verifiability and end-to-end traceability.

\begin{figure}[ht]
    \centering
    \includegraphics[width=0.95\linewidth]{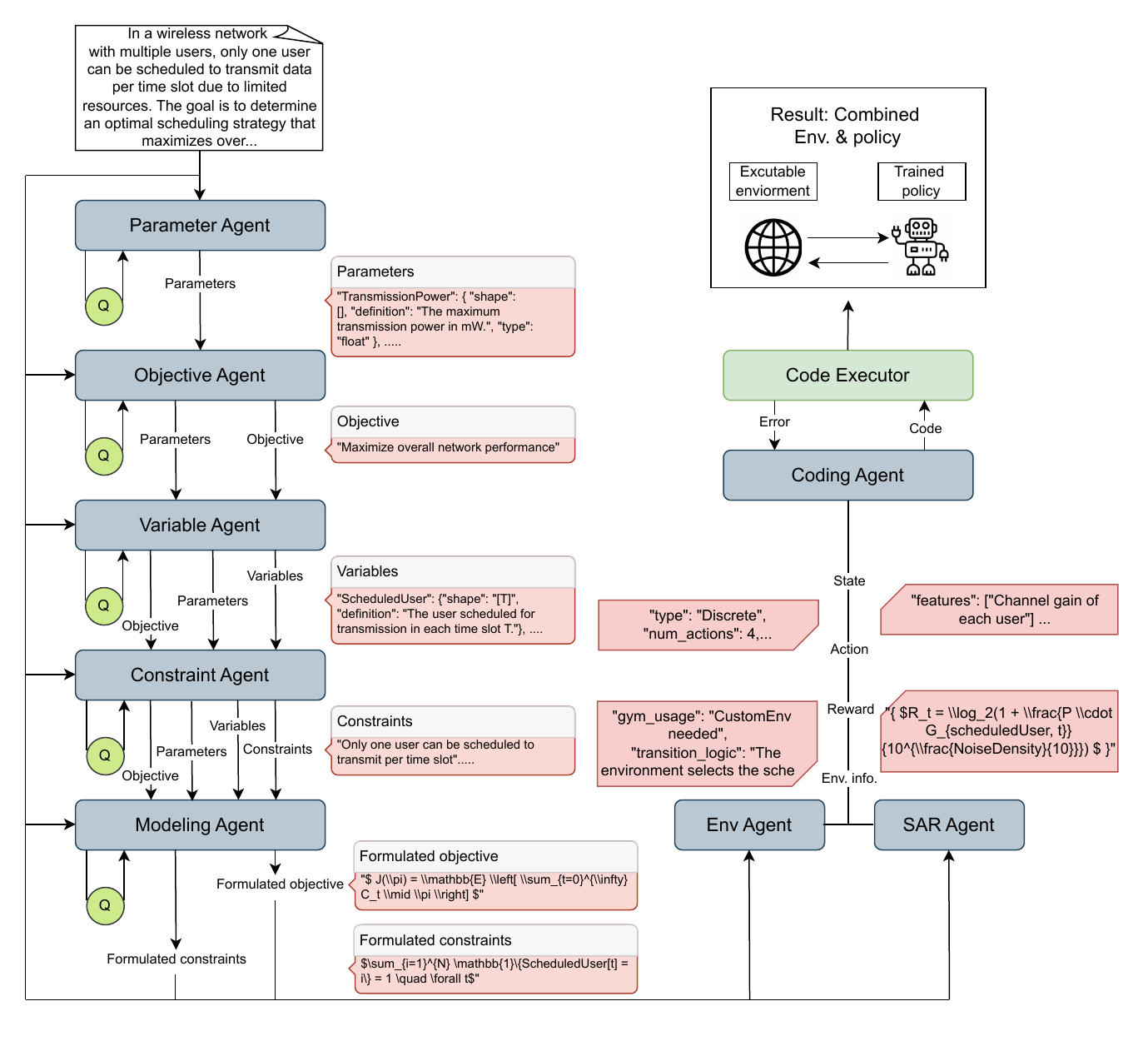}
    \caption{A use case of A-LAMP for a wireless network scheduling problem. Gray rectangles represent the specialized LLM agents, red rectangles denote the intermediate outputs produced at each agent, and green circles marked with “Q” indicate the error correction module.}
    \label{fig:polia_flow}
    \vskip -0.2in
\end{figure}

The A-LAMP process begins by entering a free-form natural language description of the target task. For example, consider the wireless network scheduling problem, whose description can be found in Appendix~\ref{sec:description}. First, in the abstract idea phase, a parameter agent identifies the key parameters that define the scope of the task (e.g., number of users). Then, an objective agent extracts a clear goal statement of the task from the natural language description, as considering the key parameters as the context for the task (e.g., maximizing system throughput).
Given both the parameters and objectives, a variable agent identifies the decision variables that can be controlled (e.g., scheduled users), and the system variables that capture environment dynamics (e.g., channel gains of users).
Based on the parameters and variables, a constraint agent encodes the feasibility rules that must be satisfied (e.g., maximum number of scheduled users).
These agents mirror the intuitive reasoning process of understanding the scope of the task, defining its goal, and determining the available control options.

The formulation phase is handled by a modeling agent and a state-action-reward (SAR) agent, which converts the objective and constraints into a formal MDP using the previously extracted parameters and decision variables. 
Concretely, the modeling agent translates the extracted objective into a standard MDP formulation based on the identified variables and parameters: $\max_{\pi:\mathcal{S}\rightarrow\mathcal{A}}\; J(\pi)$, where $\pi$ is the policy, $\mathcal{S}$ is the state space, $\mathcal{A}$ is the action space, and $J(\pi)$ is an expected cumulative reward by policy $\pi$.
%$\max_{\pi}\; J(\pi) = \mathbb{E}_{\tau \sim \pi}\!\left[\sum_{t=0}^{T-1} r_t(s_t,a_t)\right],$
%where $r_t$ is instantiated from the extracted goal (e.g., throughput reward), while $s_t$ and $a_t$ are composed of the identified variables and domain parameters. 
%
In addition to the objective, the modeling agent also formalizes the extracted constraints as explicit mathematical forms to guarantee them respected in the environment and policy.
Building on this formulation, the SAR agent explicitly defines states (e.g., vector of channel gains), actions (e.g., scheduled users), and rewards (e.g., system throughput).
This integration preserves internal consistency before moving to the coding phase.

In the first two phases, the agents primarily reason from long inputs to short extracts, in which errors are more difficult to detect.
To address this issue, an error correction module introduced in \citep{ahmaditeshnizi2024optimus} is attached to the relevant agents.
Specifically, the module instructs each agent to check its own extraction. The agent assigns itself a self-confidence score. If the score falls below a certain threshold, the agent reexamines its output. If ambiguity persists, the agent may issue a concise clarification request to a human expert.
%
%We limit this to the first two phases because their tasks are long-input$\rightarrow$short-extract reasoning, where errors are harder to surface; in contrast, the coding phase externalizes reasoning as executable code, logs and tests that can be verified directly.

In the coding phase, an environment agent and a coding agent collaborate to generate an executable RL code that incorporates both the simulated environment and RL algorithms. Then, a code executor uses the code to train a policy for the target task.
The environment agent defines the environment dynamics, including the environmental rules (e.g., a channel model with path loss and fading), and termination conditions (e.g., a time horizon of $T$), both of which are realized in the code.
Based on the dynamics and the SAR structure, the coding agent writes runnable Gym-style code and implements environment classes, reset and step functions, and RL training loops.
In cases where run-time errors or inconsistencies arise, a feedback loop between the coding agent and the code executor automatically performs debugging and correction. Through this loop, A-LAMP bridges natural language and deployable RL policies with minimal human intervention.
%, maintaining interpretability while reducing the modeling effort required of human experts.

\section{Experiments}

\subsection{Experimental Setup}
We evaluate the performance of A-LAMP on five diverse RL tasks, spanning a spectrum of complexity from standard control tasks to domain-specific optimization tasks.
In experiments, each task is specified solely through a free-form natural language description. The more details of each description can be found in the Appendix \ref{sec:description}.
% {\color{red}[consistency 맞춰서 다시 작성]}
\begin{itemize}
    \item \textbf{Cart-pole}, \textbf{Mountain-car}: Classic control tasks with well-known MDP structures and existing Gym environments, requiring no custom environment generation.
    \item \textbf{Wireless}: A multi-user resource allocation task with domain-specific formulations in wireless communications (e.g.Shannon capacity).
    \item \textbf{Drone-delivery (Drone-del.)}: A $50\times50$ grid world-based task involving package delivery under energy constraints, requiring explicit planning.
    \item \textbf{Inventory-management (Inv.-mgmt.)}: A retail inventory optimization task, where demand follows a Poisson process, requiring cost-aware replenishment decisions.
    % \item \textbf{Inventory-management (Inv.-mgmt.)}: In retail systems where demand is modeled as a Poisson process, the problem is to determine the optimal replenishment quantity for each item at every decision epoch, taking into account ordering, holding, and penalty costs associated.
\end{itemize}

We evaluate the performance of A-LAMP to generate a valid MDP formulation and an executable policy.
To this end, we define three following evaluation criteria:
% {\color{red}[다시 작성]}
% \begin{itemize}
%     \item \textbf{Modeling success rate}: Whether the extracted MDP components (state, action, reward) are logically correct and complete
%     \item \textbf{Coding success rate}: Whether the generated code executes without syntax errors in a standard Python environment.
%     \item \textbf{Policy generation success rate}: Whether RL training converges to a reward-maximizing policy, and whether the trained policy satisfies the intended task objectives. (only defined when coding success is achieved)
% \end{itemize}
\begin{itemize}
    \item \textbf{Modeling success rate}: The proportion of trials, where the extracted MDP components (state, action, reward) are logically correct and complete. Formally, it is defined as the number of trials with successful modeling divided by the total number of trials.
    \item \textbf{Coding success rate}: The proportion of trials, where the generated code executes without syntax errors in a standard Python environment. It is defined as the number of trials with successful code executions divided by the total number of trials.
    \item \textbf{Policy generation success rate}: The proportion of trials, where RL training converges to a reward-maximizing policy and the trained policy satisfies the intended task objectives. This metric is computed as the number of trials with successful policy generation divided by the total number of trials.
\end{itemize}

%

% For experimental automation, the error correction module described in Section~\ref{sec:method} was disabled, ensuring a fully autonomous pipeline without human intervention. 
We compare four methods: the A-LAMP framework with GPT-4o, the Light A-LAMP framework with Gemma3-27B, and single-model baselines using GPT-4o and Gemma3-27B. This setup allows us to examine how the model size influences the complexity of tasks that can be handled and the resulting performance.
The partial example of the prompts of A-LAMP is provided in Appendix \ref{sec:prompts}.
For policy generation, we guide the methods to adopt Deep Q-Networks (DQN) as the RL algorithm. DQN is a standard and robust choice for problems with discrete action spaces, which all of our tasks exhibit. This ensures a consistent backbone so that performance differences reflect modeling and coding ability rather than the RL algorithm itself. 

For single-model baselines, we request a Python-based training code to solve the task, providing an identical task description. Since an intermediate MDP modeling result is inaccessible in these cases, the success of MDP modeling is manually assessed by verifying the consistency of the extracted state, action space, and reward definitions within the generated code.

% \subsection{Overall Performance}
% \label{sec:performance}
% \begin{table}[ht]
%   \caption{Evaluation result of A-LAMP, Light A-LAMP, Gemma3-27B, and GPT-4o across five benchmark tasks. ``M'', ``C'', and ``P'' denote modeling, coding, and policy generation success rates, evaluated over 20 trials.}
%   \label{tab:poliacomparison}
%   \centering
%   \resizebox{\textwidth}{!}{
%     \begin{tabular}{lccccc}
%       \toprule
%       \multicolumn{1}{c}{Task} & A-LAMP (M/C/P) & A-LAMP (w/o EC.) (M/C/P) & Light A-LAMP (M/C/P) & Gemma3-27B (M/C/P) & GPT-4o (M/C/P) \\
%       \midrule
%       Cart-pole   & -         & \textbf{1.00} / \textbf{0.95} / \textbf{0.95} & \textbf{1.00} / 0.85 / 0.45 & \textbf{1.00} / 0.60 / 0.35 & \textbf{1.00} / 0.75 / 0.45 \\
%       Mountain-car & -  & \textbf{1.00} / \textbf{1.00} / \textbf{0.75} & 0.95 / 0.70 / 0.55 & \textbf{1.00} / 0.35 / 0.30 & \textbf{1.00} / \textbf{1.00} / 0.40 \\
%       Wireless  &  1.00/ 1.00 / 0.45  & 0.90 / 0.80 / \textbf{0.40} & \textbf{0.95} / 0.60 / 0.15  & 0.55 / 0.65 / 0.05 & 0.80 / \textbf{0.90} / 0.20 \\
%       Drone-del.  &  0.80/ 0.95 / 0.45  & \textbf{0.65} / \textbf{0.75} / \textbf{0.30} & 0.55 / 0.50 / 0.15  & 0.40 / 0.05 / 0.00 & 0.35 / 0.55 / 0.10 \\
%       Inv.-mgmt.    &  1.00/ 0.55 / 0.30 & \textbf{1.00} / \textbf{0.40} / \textbf{0.20} & 0.85 / 0.25 / 0.05  & 0.60 / 0.00 / 0.00 & 0.65 / 0.05 / 0.05 \\
%       \bottomrule
%     \end{tabular}
%   }
% \end{table}

\subsection{Overall Performance}
\label{sec:performance}
\begin{table}[ht]
  \caption{Evaluation result of A-LAMP, A-LAMP without error correction (EC), Light A-LAMP, Gemma3-27B, and GPT-4o across five benchmark tasks. 
  All results are reported as triplets in the order of modeling, coding, and policy generation success rates, which are evaluated over 20 trials.}
  %A-LAMP was additionally tested with error correction (EC) module enabled only for selected three tasks (wireless, drone-delivery, and inventory-management).}
  \label{tab:poliacomparison}
  \centering
  \resizebox{\textwidth}{!}{
    \begin{tabular}{lcccccc}
      \toprule
      \multicolumn{1}{c}{Task} & \multicolumn{1}{c}{A-LAMP} & \multicolumn{1}{c}{A-LAMP w/o EC} & \multicolumn{1}{c}{Light A-LAMP} & \multicolumn{1}{c}{Gemma3-27B} & \multicolumn{1}{c}{GPT-4o} \\
  %    \cmidrule(lr){2-3}
  %     & (w/ EC.) (M/C/P) & (M/C/P) & (M/C/P) & (M/C/P) & (M/C/P) \\
      \midrule
      Cart-pole    & -   & \textbf{1.00} / \textbf{0.95} / \textbf{0.95} & \textbf{1.00} / 0.85 / 0.45 & \textbf{1.00} / 0.60 / 0.35 & \textbf{1.00} / 0.75 / 0.45 \\
      Mountain-car & -   & \textbf{1.00} / \textbf{1.00} / \textbf{0.75} & 0.95 / 0.70 / 0.55 & \textbf{1.00} / 0.35 / 0.30 & \textbf{1.00} / \textbf{1.00} / 0.40 \\
      Wireless     & \textbf{1.00} / \textbf{1.00} / \textbf{0.45} & 0.90 / 0.80 / 0.40 & 0.95 / 0.60 / 0.15  & 0.55 / 0.65 / 0.05 & 0.80 / 0.90 / 0.20 \\
      Drone-del.   & \textbf{0.80} / \textbf{0.95} / \textbf{0.45} & 0.65 / 0.75 / 0.30 & 0.55 / 0.50 / 0.15  & 0.40 / 0.05 / 0.00 & 0.35 / 0.55 / 0.10 \\
      Inv.-mgmt.   & \textbf{1.00} / \textbf{0.55} / \textbf{0.30} & \textbf{1.00} / 0.40 / 0.20 & 0.85 / 0.25 / 0.05  & 0.60 / 0.00 / 0.00 & 0.65 / 0.05 / 0.05 \\
      \bottomrule
    \end{tabular}
  }
\end{table}

% Table~\ref{tab:poliacomparison} summarizes the success rates of
% % M, C, and P
% three evaluation criteria (M, C, and P)
% for each benchmark task and model. Among them, the policy generation success rate (denoted as P) is the most critical metric, and it clearly demonstrates the strength of our approach. A-LAMP consistently achieves the highest policy generation success rate across all tasks, and its advantage is especially pronounced in custom environments such as drone-del., Inv.-mgmt., and wireless, where it attains nearly twice policy generation success rate of single-model baselines. This confirms that A-LAMP enhances the efficiency of automated end-to-end policy generation.

Table~\ref{tab:poliacomparison} summarizes the success rates of three evaluation criteria for each benchmark task and method. Among them, the policy generation success rate is the most critical metric, and it clearly demonstrates the strength of our approach. 
We provide the results of A-LAMP with an error correction (EC) module only for selected three tasks (wireless, drone-delivery, and inventory-management).
Even without the error correction (EC) module, fully automated A-LAMP consistently achieves the highest policy generation success rate across all tasks, with its advantage especially pronounced in tasks which need to generate custom environments (e.g., drone-delivery, inventory-management, and wireless). In these tasks, A-LAMP attains nearly twice the policy generation success rate of single-model methods, confirming its ability increase the success rate of automated end-to-end policy generation.

Beyond this, Light A-LAMP with the much smaller Gemma3-27B achieves substantially higher policy generation success rate than Gemma3-27B alone and even approaches the performance of GPT-4o. Considering the large gap in parameter scale, this shows that the performance gain derives from A-LAMP rather than model size. These results demonstrate that our framework extends beyond the range of RL problems that a single LLM model can handle, enabling reliable policy generation even in more complex domains.

Furthermore, the strong performance achieved without the error correction module highlights that the structural decomposition of the proposed framework alone produces substantial gains. With the error correction module selectively enabled, performance in three tasks, particularly those requiring custom environment generation (e.g., drone-delivery, inventory-management, and wireless) improved further, demonstrating that minimal human intervention can amplify the benefits of our framework under difficult tasks.

% Taken together, the findings suggest that even RL problems that are prohibitively difficult for single-model prompting can be tackled effectively through our agentic decomposition. Moreover, the current evaluation excludes human feedback and the built-in error correction module, yet A-LAMP already exhibits a clear performance advantage. 

% Incorporating these additional mechanisms is likely to further widen the gap, reinforcing the promise of our framework for scalable and reliable RL automation.

\subsection{Understanding Performance Gains: A-LAMP vs. a Single LLM}
\label{sec:failure}

In this section, we explain the performance gains observed in Section~\ref{sec:performance} by analyzing failure cases rather than directly interpreting overall success rates. In all methods and tasks, a total of 460 trials were conducted, providing an empirical basis for our analysis. Since a success case for policy generation necessarily requires that modeling, coding, and training are all successful simultaneously, analyzing only the success case makes it difficult to unravel which criterion contributed the most to the performance gain. In contrast, failure cases allow us to observe how each criterion improves independently and to quantify where A-LAMP provides the most benefit.
For clarity, we denote MDP modeling, coding, and policy generation by M, C, and P, respectively. Success and failure are represented by $\circ$ and $\times$. For instance, M$\circ$ indicates successful MDP modeling, whereas M$\times$ indicates failure. The same notation applies to C and P. The basis for this analysis comes from the distribution of failure cases, which is provided in Appendix~\ref{sec:fail_append}.

\begin{figure}[ht]
    \begin{center}
    \subfigure[Distribution of A-LAMP w/o EC.]{
        \includegraphics[width=0.45\textwidth]{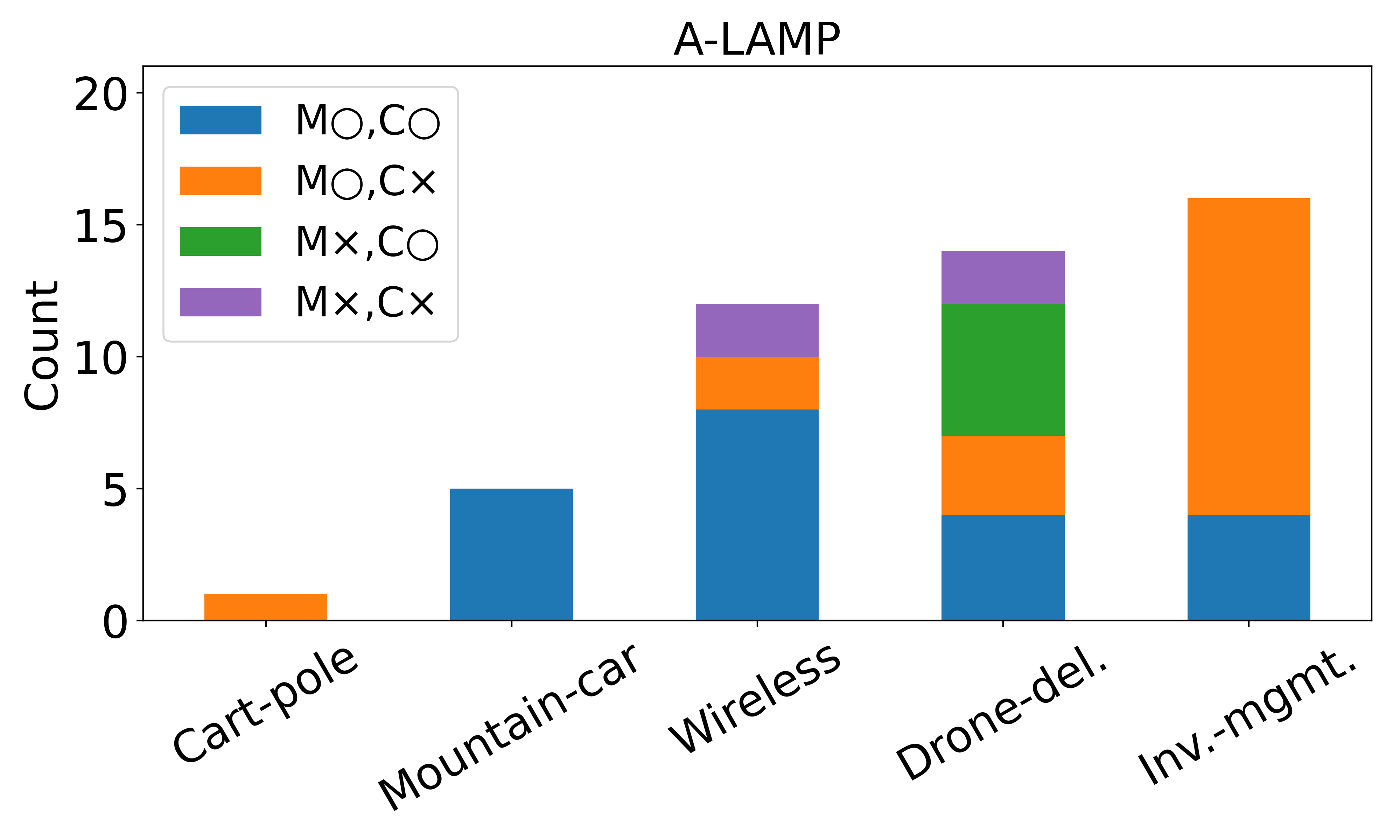}
        \label{fig:A_LAMP_dis}
    }
    \hfil
    \subfigure[Distribution of a single GPT-4o.]{
        \centering
        \includegraphics[width=0.45\textwidth]{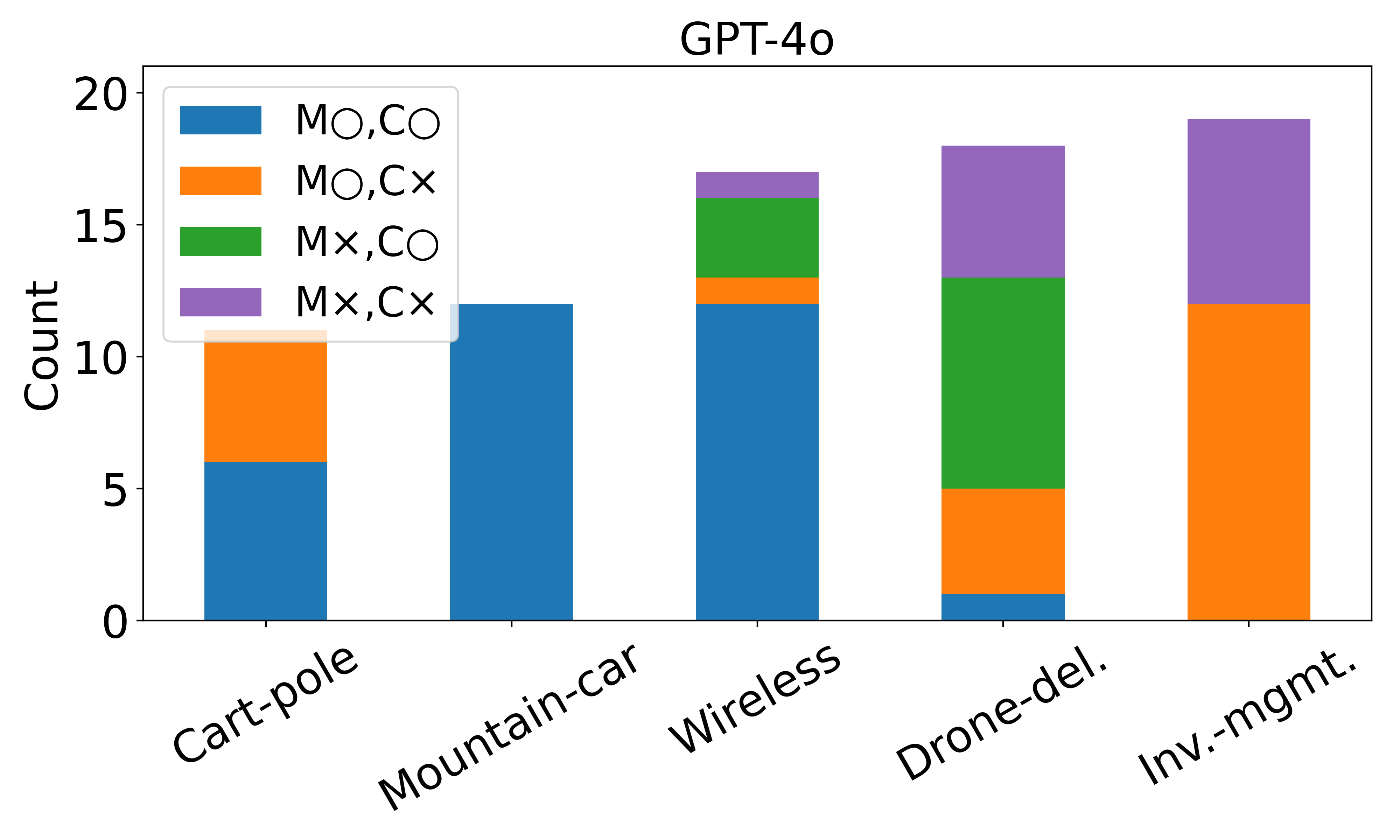}
        \label{fig:single_GPT_dis}
    }
    \caption{Failure distributions in the cases of $\text{P}\times$}
    \label{fig:gpt_LAMP_failure}
    \end{center}
    \vskip -0.1in
\end{figure}

This motivates a closer look at how failure cases are distributed between different criteria. We here consider A-LAMP without EC to remove the impact of error correction.
As illustrated in Figure~\ref{fig:gpt_LAMP_failure}, the cases of P$\times$ can be decomposed into the cases of M and C results, allowing a direct comparison between A-LAMP w/o EC and a single GPT-4o. First, the portion of P$\times$ where C succeeds without valid M (M$\times$,C$\circ$) is greatly reduced under A-LAMP. In the single GPT-4o, this outcome appeared in wireless and drone-delivery tasks, where the code was syntactically correct but semantically trivial and disconnected from the task. With A-LAMP, this spurious coding success is entirely removed in wireless and reduced by 37.5\% in drone-delivery, indicating improved alignment between M and C. 
Second, the complete failure mode (M$\times$,C$\times$) is also reduced under A-LAMP, especially in the more complex tasks. In drone-delivery, this outcome decreases by 60\%, and in inventory-management, it is eliminated entirely (100\%). This demonstrates that A-LAMP strengthens M reliability by reducing instances where neither M nor C succeed. 
Third, A-LAMP substantially improves training stability. In cart-pole and mountain-car, which already had relatively high baseline success, a large fraction of runs that previously failed at the training stage (91\% and 58\%) are successfully converted into policies under A-LAMP. This demonstrates that decomposition not only improves M and C but also enhances the robustness of the final training phase.

Overall, the results indicate that A-LAMP improves P not only as shown in Section~\ref{sec:performance}, but also through insights gained from the failure analysis: (1) by boosting M and eliminating meaningless coding-only outcomes, (2) by reducing complete failures to strengthen M reliability, and (3) by stabilizing training dynamics to turn more $(\text{M}\circ,\text{C}\circ)$ attempts into successful P.

\subsection{Policy Training and Evaluation on Wireless Task}
\label{sec:case_study_ori}
Beyond the improvement in policy generation success rate achieved by A-LAMP, we verify whether the generated policies operate as intended in tasks. To isolate coding/executability from domain-specific modeling and to enable straightforward evaluation, the wireless task is designed so that at each time step, a single user is scheduled and the reward is given as the instantaneous sum-rate. There are no inter-temporal couplings (e.g., time-average quality of service). Under this formulation, selecting the user with the highest instantaneous rate is optimal; hence a greedy scheduler is the optimal policy.

\begin{figure}[ht]
\vspace{-1em}
    \begin{center}
    \subfigure[Training progress.]{
        \includegraphics[width=0.45\textwidth]{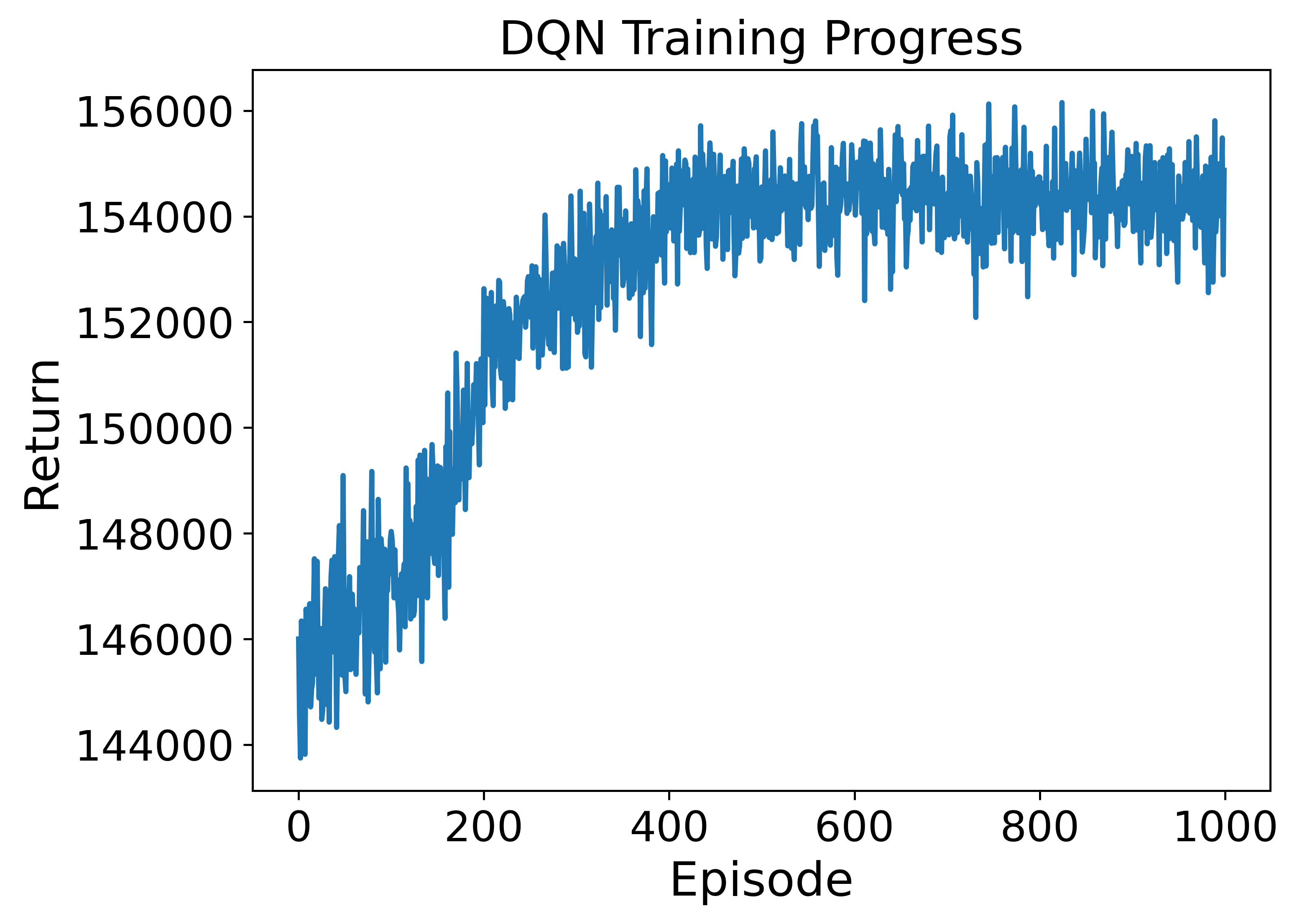}
        \label{fig:wireless_train}
    }
    \hfil
    \subfigure[Policy evaluation: DQN vs. Greedy baseline.]{
        \centering
        \includegraphics[width=0.45\textwidth]{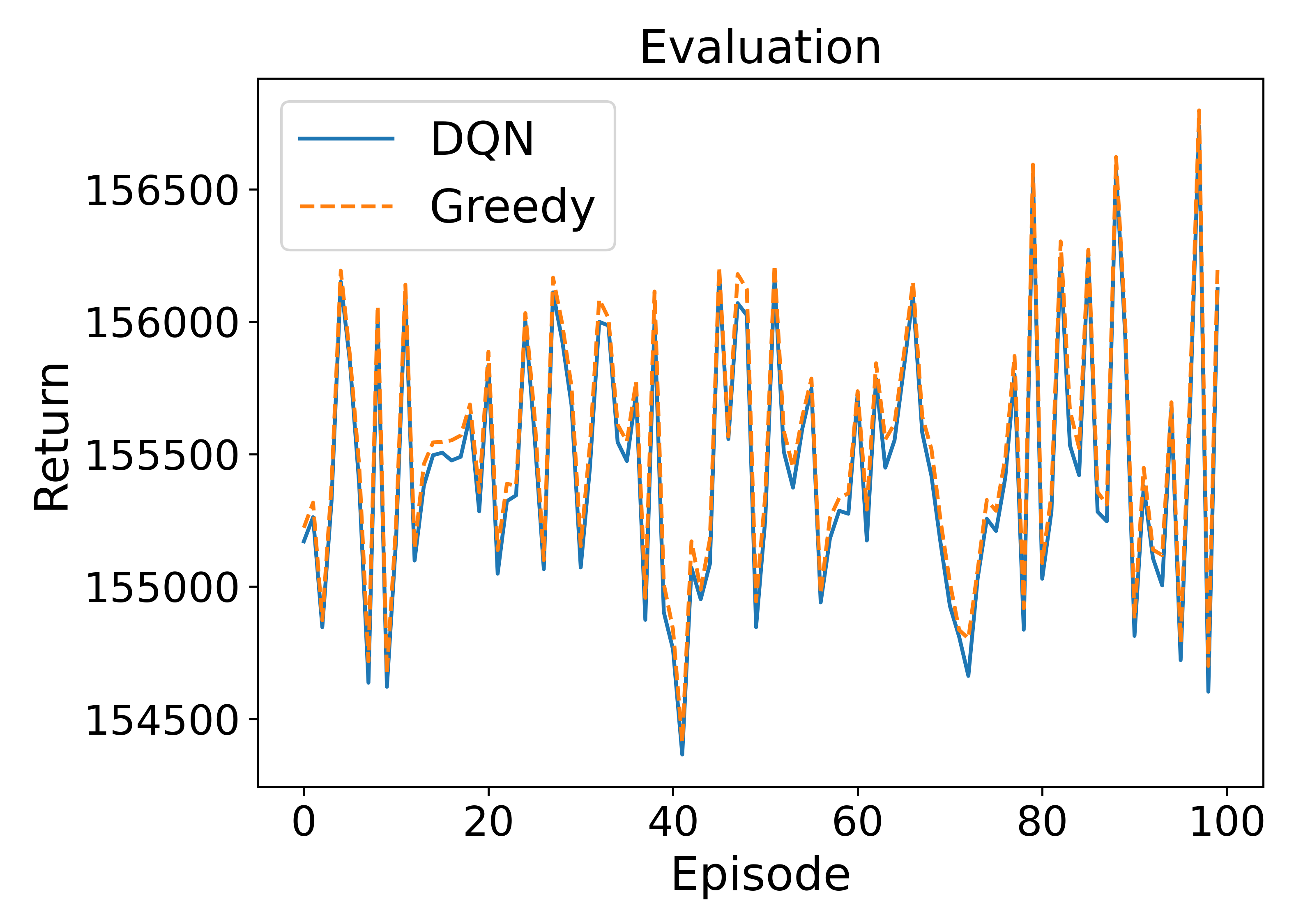}
        \label{fig:wireless_eval}
    }
    \caption{The return of DQN generated by A-LAMP in the training and evaluation stages.}
    %\caption{A-LAMP enables stable DQN training and generates a learned policy with performance close to the optimal greedy scheduler.}
    \label{fig:wireless_case}
    \end{center}
    \vskip -0.1in
\end{figure}

% As shown in Table~\ref{tab:poliacomparison}, A-LAMP achieves nearly twice the policy generation success rate of GPT-4o in the wireless environment.

Figure~\ref{fig:wireless_case} presents the DQN training and evaluation results generated by A-LAMP. In Figure~\ref{fig:wireless_train}, the training curve rises quickly and stabilizes, confirming that the agent acquires a well-formed and logically consistent policy. In Figure~\ref{fig:wireless_eval}, the learned DQN closely approaches the performance and stability of the greedy scheduler. Since the greedy scheduler is optimal, this proximity demonstrates that the policy generated by A-LAMP not only executes correctly but also preserves the intended optimality structure of the task. The additional case study of other tasks can be found in the Appendix ~\ref{sec:case_study}.\looseness=-1

\section{Conclusion}

We proposed A-LAMP, a modular multi-agent LLM framework that automates MDP modeling and policy generation from free-form natural language descriptions. The framework decomposes the extraction of parameters, objectives, variables, and constraints, followed by MDP formulation and code generation. This process ensures semantic alignment from task description to executable policy, thereby reducing expert effort.
Through extensive experiments across classic control and domain-specific tasks, we demonstrated that A-LAMP consistently outperforms single-model baselines. Its improvements are clearly shown not only in overall policy generation success but also in each specific stage, enhancing modeling reliability and stabilizing training dynamics. A case study further confirmed that the generated policies preserve task-level optimality, validating both executability and correctness. These findings highlight that structured multi-agent decomposition is the key to enabling reliable and scalable RL automation.

For future work, we identify several directions. Although the current decomposition ensures correctness in MDP modeling, the coding stage remains fragile; syntactic or structural mismatches can still hinder executability. One way to improve robustness and reduce debugging costs is to extend the framework with finer-grained coding agents, such as dedicated modules for environment construction and training loop generation. Moreover, incorporating adaptive mechanisms for hyperparameter tuning and validation, as well as domain-informed priors and structured knowledge integration, will further generalize A-LAMP to complex domains.

\section*{Acknowledgements}

This work was supported in part by the National Research Foundation of Korea (NRF) grant funded by the Korea government (MSIT) (RS-2025-24523498), in part by the Institute of Information \& Communications Technology Planning \& Evaluation(IITP)-ITRC(Information Technology Research Center) grant funded by the Korea government (MSIT) (IITP-2025-RS-2021-II211816), and in part by the Technology Innovation Program (RS-2022-00154678, Development of Intelligent Sensor Platform Technology for Connected Sensor) funded by the Ministry of Trade, Industry \& Energy(MOTIE, Korea).
We thank all reviewers for their comments and suggestions.

\bibliographystyle{unsrtnat}
\bibliography{example_paper_(2)}

% \section*{References}

% References follow the acknowledgments in the camera-ready paper. Use unnumbered first-level heading for
% the references. Any choice of citation style is acceptable as long as you are
% consistent. It is permissible to reduce the font size to \verb+small+ (9 point)
% when listing the references.
% Note that the Reference section does not count towards the page limit.
% \medskip

% {
% \small

% [1] Alexander, J.A.\ \& Mozer, M.C.\ (1995) Template-based algorithms for
% connectionist rule extraction. In G.\ Tesauro, D.S.\ Touretzky and T.K.\ Leen
% (eds.), {\it Advances in Neural Information Processing Systems 7},
% pp.\ 609--616. Cambridge, MA: MIT Press.

% [2] Bower, J.M.\ \& Beeman, D.\ (1995) {\it The Book of GENESIS: Exploring
%   Realistic Neural Models with the GEneral NEural SImulation System.}  New York:
% TELOS/Springer--Verlag.

% [3] Hasselmo, M.E., Schnell, E.\ \& Barkai, E.\ (1995) Dynamics of learning and
% recall at excitatory recurrent synapses and cholinergic modulation in rat
% hippocampal region CA3. {\it Journal of Neuroscience} {\bf 15}(7):5249-5262.
% }

%%%%%%%%%%%%%%%%%%%%%%%%%%%%%%%%%%%%%%%%%%%%%%%%%%%%%%%%%%%%
\clearpage
\appendix

\section{Policy Generation Failure Distribution Analysis}
\label{sec:fail_append}
\begin{table}[ht]
  \caption{Normalized result distribution per task. 
Within each task, P$\circ$ and P$\times$ groups are normalized separately to sum to 1.0, where $\circ$ denotes success and $\times$ denotes failure.}
  \label{tab:env_outcome_grouped_norm}
  \centering
  % \resizebox{\textwidth}{!}{
    \begin{tabular}{l|cc|cccc}
      \toprule
       & \multicolumn{2}{c|}{P$\circ$} & \multicolumn{4}{c}{P$\times$} \\
      Task & M$\circ$ & M$\times$ & M$\circ$,C$\circ$ & M$\circ$,C$\times$ & M$\times$,C$\circ$ & M$\times$,C$\times$ \\
      \midrule
      Cart-pole   & 1.00 & 0.00 & 0.53 & 0.47 & 0.00 & 0.00 \\
      Mountain-car   & 1.00 & 0.00 & 0.50 & 0.48 & 0.03 & 0.00 \\
      Wireless   & 1.00 & 0.00 & 0.54 & 0.22 & 0.14 & 0.11 \\
      Drone-del.     & 1.00 & 0.00 & 0.13 & 0.28 & 0.25 & 0.35 \\
      Inv.-mgmt.  & 1.00 & 0.00 & 0.09 & 0.68 & 0.01 & 0.22 \\
      \bottomrule
    \end{tabular}
  % }ㅊ
\end{table}

Table~\ref{tab:env_outcome_grouped_norm} summarizes the distribution of policy generation outcomes across modeling (M), coding (C), and training (P) for all tasks. Each row aggregates the results over all methods and trials for the corresponding task, and shows how failures decompose into distinct failure type (e.g., M$\circ$,C$\circ$,P$\times$ vs. M$\times$,C$\circ$,P$\times$). This aggregation enables independent observation of which stage forms the bottleneck. These distributions form the empirical basis for the failure analysis in Section~\ref{sec:failure}, where we highlight that A-LAMP not only improves M and C reliability but also reduces spurious coding-only cases and stabilizes training dynamics.

% In this section, we aim to explain the performance gains observed in Section~\ref{sec:performance} by analyzing the underlying causes of P failures. Rather than directly interpreting overall success rates, we examine failure cases to reveal why A-LAMP achieves higher P than a single LLM. To understand this choice, it is important to clarify the causal dependencies among M, C, and P. Table~\ref{tab:env_outcome_grouped_norm} shows that P$\circ$ never occurs without prior M and C success ($\text{M}\times,\text{P}\circ=0$ and $\text{M}\circ,\text{C}\times,\text{P}\circ=0$), confirming that $\text{P} \subset (\text{M} \cap \text{C})$. However, each task has nontrivial mass in $\text{M}\circ,\text{C}\circ,\text{P}\times$ (e.g., 53\% in cart-pole, 50\% in mountain-car, 54\% in wireless), indicating that $(\text{M} \cap \text{C}) \not\subset \text{P}$ and that training stability constitutes an additional bottleneck for successful P. Therefore, the bottlenecks to successful P can be summarized into three sequential stages: M first, then C, and finally training stability for P.

\section{Natural Language Descriptions of Benchmark Tasks}
\label{sec:description}

\begin{tcolorbox}[title=Cart-Pole]
There is a cart that can move left and right and a pole is attached on a cart.
The goal is to determine how to move the cart so that the pole remains upright for as long as possible. 
I will solve this problem using reinforcement learning.
\end{tcolorbox}

\begin{tcolorbox}[title=Mountain-Car]
A car is placed between two hills and must build enough momentum to reach the top of the right hill. 
The agent can accelerate left, right, or stay still. The reward is -1 per time step until the goal is reached. 
I will solve this problem using reinforcement learning.
\end{tcolorbox}

\begin{tcolorbox}[title=Wireless]
In a wireless network with multiple users, only one user can be scheduled to transmit data per time slot due to limited resources.  
The goal is to determine an optimal scheduling strategy that maximizes overall network performance while adhering to system constraints.  
Each user experiences a time-varying channel influenced by signal strength, interference, and noise.  
The transmission rate depends on the selected user's channel gain, transmission power, and environmental noise, and it follows Shannon’s capacity formula.  
The system operates with a $5$ MHz bandwidth and $10,000$ mW transmission power. Noise density is set at $-106$ dBm, and the channel is affected by a path loss coefficient of $3.76$ and log-normal shadowing with a $10$ dB standard deviation.  
The environment includes $4$ users located at varying distances from the base station ($20$ m, $50$ m, $50$ m, and $80$ m), which influences their individual channel quality.  
Channel gains are normalized between $-80$ dB and $-30$ dB to reflect variations in signal strength.
\end{tcolorbox}

\begin{tcolorbox}[title=Drone-Delivery]
The drone operates in a $50 \times 50$ grid world, where it must deliver packages to multiple, randomly assigned destinations. 
Both the pickup locations and delivery targets of the packages, as well as the total number of packages, are determined randomly at the start of each episode. 
The drone begins with an initial energy level randomly generated between $100$ and $150$. 
Each movement to an adjacent cell consumes $1$ unit of energy, while each package delivery consumes $2$ units of energy. 
The drone must plan its route efficiently to complete all deliveries before its energy is depleted.
I will solve this problem using reinforcement learning.
\end{tcolorbox}

\begin{tcolorbox}[title=Inventory-Management]
The system manages inventory for $10$ different items. At each time step, it observes the current stock of each item and decides how much to order.  
Each order incurs a fixed cost and a per-unit cost, and holding inventory also results in a time-based holding cost.  
Demand for each item is random, following a Poisson distribution with a mean of $8$.  
If demand exceeds stock, lost sales occur and penalties are applied.  
The goal is to fulfill demand while maximizing long-term profit.

Each item has its own cost structure. For example:  
Item 1 has a fixed cost of \$20, unit cost \$5, holding cost \$0.50, and sells for \$12.  
Item 2: fixed \$18, unit \$4, holding \$0.40, price \$10.  
Item 3: fixed \$25, unit \$6, holding \$0.60, price \$15.  
Item 4: fixed \$22, unit \$5, holding \$0.50, price \$13.  
Item 5: fixed \$19, unit \$4.5, holding \$0.45, price \$11.  
Item 6: fixed \$21, unit \$5.2, holding \$0.52, price \$14.  
Item 7: fixed \$15, unit \$3.8, holding \$0.38, price \$9.  
Item 8: fixed \$28, unit \$6.5, holding \$0.65, price \$16.  
Item 9: fixed \$16, unit \$4, holding \$0.40, price \$10.  
Item 10: fixed \$24, unit \$5.8, holding \$0.58, price \$13.5.

I will solve this problem using reinforcement learning.
\end{tcolorbox}

\section{Additional Case Study}
\label{sec:case_study}

In this section, we provide detailed case studies that illustrate how A-LAMP translates free-form language descriptions into executable RL environments and policy. For each task, the extracted outputs correspond to successful policy generation cases among 20 trials (Table~\ref{tab:poliacomparison}). To ensure reproducibility and clarity, all case studies are organized under a unified structure: each begins with the original task description, followed by the JSON-formatted outputs specifying parameters, variables, objectives, and  constraints. The environment dynamics are then detailed, including state, action, and reward definitions, and finally, empirical results are presented through training curves and policy evaluation graphs. For visualization purposes, the extracted JSON files have been minimally edited without altering their original semantics or structure.

\subsection{Case Study: Cart-Pole}

\begin{tcolorbox}[title=Parameters (JSON)]
\begin{itemize}
  \item \textbf{CartPosition}: The position of the cart (float).
  \item \textbf{CartVelocity}: The velocity of the cart (float).
  \item \textbf{PoleAngle}: The angle of the pole from vertical (float).
  \item \textbf{PoleAngularVelocity}: The angular velocity of the pole (float).
  \item \textbf{CartMaxVelocity}: The maximum velocity of the cart (float).
  \item \textbf{PoleAngleLimit}: The maximum angle the pole can deviate from vertical (float).
\end{itemize}
\end{tcolorbox}

\begin{tcolorbox}[title=Variables (JSON)]
\begin{itemize}
  \item \textbf{CartAcceleration}: The acceleration of the cart along the horizontal axis.
\end{itemize}
\end{tcolorbox}

\begin{tcolorbox}[title=Objective (JSON)]
The goal is to maximize the duration for which the pole remains upright.  
Formally: 
\begin{equation}
  J(\pi) = \mathbb{E} \left[ \sum_{t=0}^{\infty} \gamma^t R_t \;\middle|\; \pi \right]
\end{equation}
\end{tcolorbox}

\begin{tcolorbox}[title=Constraints (JSON)]
\begin{itemize}
  \item PoleAngle must be within a range to keep the pole from falling.
  \begin{equation}
    -\theta_{\text{max}} \leq \text{PoleAngle} \leq \theta_{\text{max}}
  \end{equation}

  \item CartPosition must remain within track limits. 
    \begin{equation}
    \text{CartPosition}_{\min} \leq \text{CartPosition} \leq \text{CartPosition}_{\max}
  \end{equation}
  % ($CartPosition_{\min} \leq CartPosition \leq CartPosition_{\max}$).
  \item CartVelocity should not exceed maximum allowable velocity.
  \begin{equation}
    |\text{CartVelocity}| \leq \text{MaxCartVelocity}
  \end{equation}
  % ($|CartVelocity| \leq MaxCartVelocity$).
  \item PoleAngularVelocity should remain within stable range. 
  \begin{equation}
    -\text{Range} \leq \text{PoleAngularVelocity} \leq \text{Range}
  \end{equation}
  % ($-Range \leq PoleAngularVelocity \leq Range$).
  \item CartPosition should always be non-negative. 
  \begin{equation}
    \text{CartPosition} \geq 0
  \end{equation}
  % ($CartPosition \geq 0$).
  \item CartVelocity should be non-negative. 
  \begin{equation}
    \text{CartVelocity} \geq 0
  \end{equation}
  % ($CartVelocity \geq 0$).
  \item PoleAngle must allow for instantaneous return to vertical. 
  \begin{equation}
    |\text{PoleAngle}| \leq \epsilon \;\;\implies\;\; \text{CartAcceleration} = -k \cdot \text{PoleAngularVelocity}
  \end{equation}
  % ($|PoleAngle| \leq \epsilon \implies CartAcceleration = -k \cdot PoleAngularVelocity$).
\end{itemize}
\end{tcolorbox}

\begin{tcolorbox}[title=Environment (JSON)]
This environment follows the standard Gym implementation.  
Usage: \texttt{CartPole-v1}.  
Transition logic: Adjust cart position and velocity based on applied acceleration; update pole angle and angular velocity using physics equations; ensure constraints are satisfied at each step.
\end{tcolorbox}

\begin{tcolorbox}[title=State (JSON)]
The state space represents the environment’s status at a single time step.  
Variables: PoleAngle, CartPosition, CartVelocity, PoleAngularVelocity.  
Shape: [4,].
\end{tcolorbox}

\begin{tcolorbox}[title=Action (JSON)]
The action space is defined by the force applied to the cart.  
Variables: force.  
Shape: [1].  
Type: discrete.
\end{tcolorbox}

\begin{tcolorbox}[title=Reward (JSON)]
The agent receives a reward of 1 if the pole remains upright and the cart stays within bounds; otherwise, 0.  
Formally:

\begin{equation}
R_t = 
\begin{cases}
1, & \text{if } |\text{PoleAngle}_t| \leq \theta_{\text{max}} \ \text{and } |\text{CartPosition}_t| \leq \text{CartPosition}_{\max} \\
0, & \text{otherwise}
\end{cases}
\end{equation}

where $R_t$: Reward at time $t$; $|\text{PoleAngle}_t|$: pole’s tilt; $\theta_{\text{max}}$: maximum allowed tilt; $|\text{CartPosition}_t|$: cart displacement; $\text{CartPosition}_{\max}$: maximum cart displacement.

\end{tcolorbox}

Based on the extracted JSON specification, we trained the generated RL code, and the resulting training loss curve and policy evaluation outcomes are shown below.

\begin{figure}[ht]
    \begin{center}
    \subfigure[Training progress of DQN by A-LAMP]{
        \includegraphics[width=0.45\textwidth]{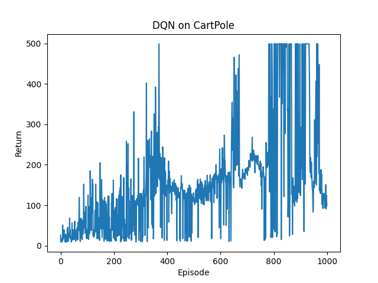}
    }
    \hfil
    \subfigure[Policy evaluation: DQN vs. Optimal baseline.]{
        \centering
        \includegraphics[width=0.45\textwidth]{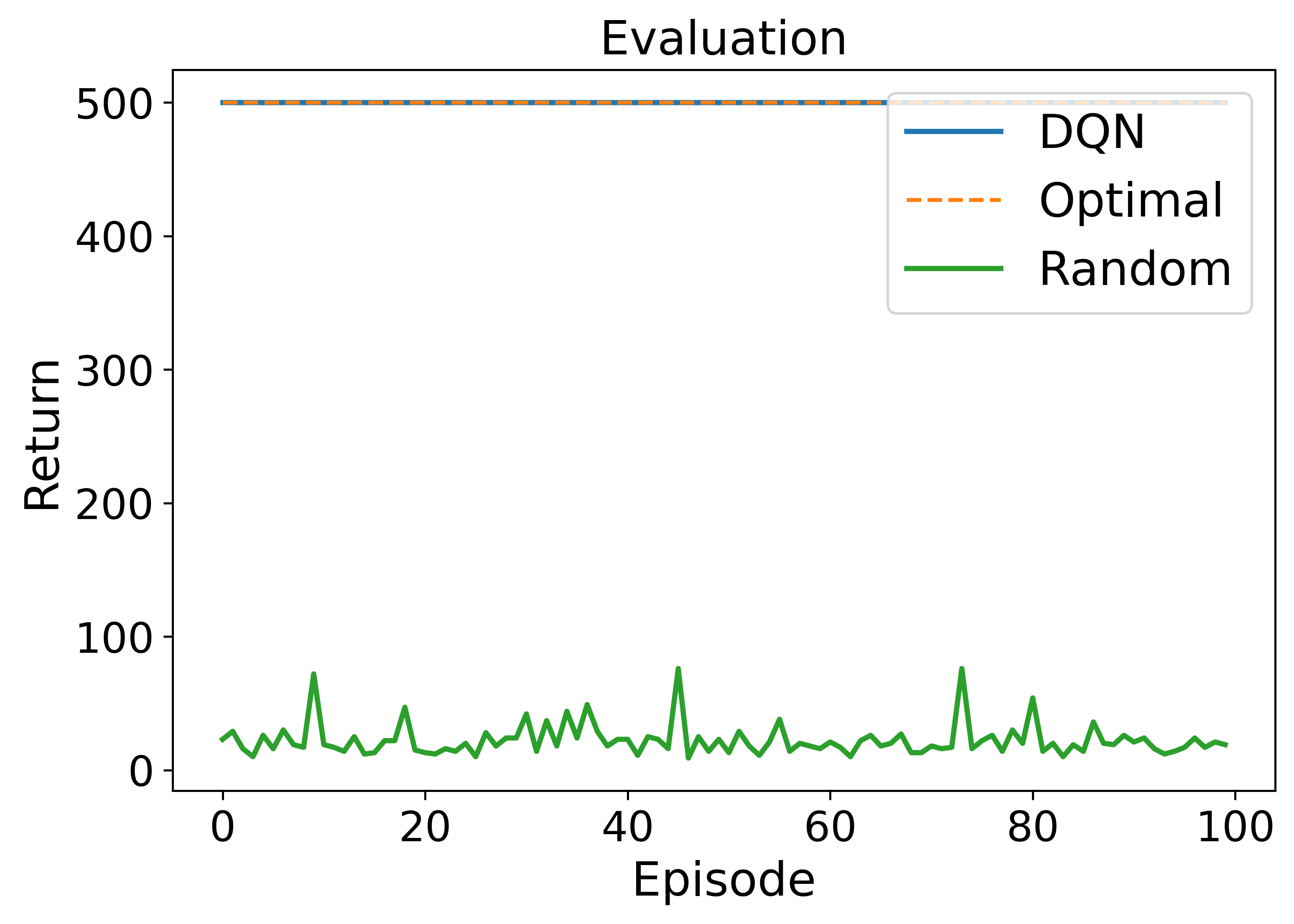}
    }
    \caption{Case study. A-LAMP successfully generates an executable environment enabling stable DQN training and competitive policy performance.}
    \end{center}
\end{figure}

\subsection{Case Study: Mountain-Car}

\begin{tcolorbox}[title=Parameters (JSON)]
\begin{itemize}
  \item \textbf{InitialReward}: The reward received at each time step before reaching the goal (float).
  \item \textbf{GoalReward}: Reward received when reaching the goal (float).
  \item \textbf{ActionSpaceSize}: Number of possible actions (3).
\end{itemize}
\end{tcolorbox}

\begin{tcolorbox}[title=Variables (JSON)]
\begin{itemize}
  \item \textbf{InitialPosition}: The initial position of the car between the two hills.
  \item \textbf{HillHeights}: The heights of the left and right hills.
  \item \textbf{HillDistances}: Distances to the hills from the initial position.
  \item \textbf{CarMass}: Mass of the car.
  \item \textbf{MaxAcceleration}: Maximum acceleration the car can achieve.
  \item \textbf{TimeStepDuration}: Duration of each time step.
  \item \textbf{Gravity}: Gravitational acceleration.
  \item \textbf{FrictionCoefficient}: Friction coefficient between car and ground.
\end{itemize}
\end{tcolorbox}

\begin{tcolorbox}[title=Objective (JSON)]
The goal is to minimize the cumulative reward (equivalent to minimizing the number of time steps) to reach the top of the right hill.  
Formally:
\begin{equation}
  J(\pi) = \mathbb{E}_{\pi} \left[ \sum_{t=0}^{T} R_t \right]
\end{equation}
\end{tcolorbox}

\begin{tcolorbox}[title=Constraints (JSON)]
\begin{itemize}
  \item The car must reach the top of the right hill to terminate the episode.
  \begin{equation}
    \text{Position} \geq \text{HillDistances}[1] + \text{HillHeights}[1]
  \end{equation}

  \item The car's position and velocity must be physically plausible.
  \begin{equation}
    |\text{V}_t| \leq \text{MaxAcceleration} \cdot \text{TimeStepDuration},
  \end{equation}
  
  \begin{equation}
    \text{P}_{t+1} = \text{P}_t + \text{V}_t \cdot \text{TimeStepDuration} + 0.5 \cdot \text{A}_t \cdot \text{TimeStepDuration}^2
  \end{equation}

  \item The car's acceleration is limited to discrete actions.
  \begin{equation}
    \{ a_{\text{left}}, a_{\text{right}}, a_{\text{still}} \}
  \end{equation}

  \item The car starts between the two hills.
  \begin{equation}
    0 \leq \text{InitialPosition} \leq \text{HillDistances}[1] + \text{HillDistances}[0]
  \end{equation}

  \item The car's position must remain within the environment.
  \begin{equation}
    0 \leq \text{Position} \leq \sum_{i=1}^{2} \text{HillDistances}[i]
  \end{equation}

  \item Time steps are discrete.
  \begin{equation}
    t \in \mathbb{Z}_{\geq 0}
  \end{equation}

  \item The reward is $-1$ until the goal is reached, implying minimization of time steps.
  \begin{equation}
    \sum_{t=0}^{T-1} -\text{InitialReward} = -T
  \end{equation}

  \item The car must build enough momentum to overcome the right hill.
  \begin{equation}
    0.5 \cdot \text{CarMass} \cdot \text{Velocity}^2 \geq \text{CarMass} \cdot \text{Gravity} \cdot \text{HillHeights}[1]
  \end{equation}
\end{itemize}
\end{tcolorbox}

\begin{tcolorbox}[title=Environment (JSON)]
This environment requires a custom Gym implementation.  
Transition logic: Based on the chosen action (left, right, still), update the car's velocity considering acceleration, time step duration, and friction.  
Calculate the new position from the updated velocity. If the car reaches the goal, terminate the episode; otherwise, apply the reward and continue.
\end{tcolorbox}

\begin{tcolorbox}[title=State (JSON)]
The state space represents the car's current situation, including its position and velocity.  
Variables: \textbf{CarPosition}, \textbf{CarVelocity}.  
Shape: [2,].
\end{tcolorbox}

\begin{tcolorbox}[title=Action (JSON)]
The action space consists of three discrete actions: accelerating left, accelerating right, or staying still.  
Variables: \textbf{Acceleration}.  
Shape: [1].  
Type: discrete.
\end{tcolorbox}

\begin{tcolorbox}[title=Reward (JSON)]
The reward function provides feedback to the agent.  
A negative reward is given for each time step, encouraging the agent to reach the goal as quickly as possible.  
A positive reward is given when the goal is reached.  
Formally:
\begin{equation}
R_t =
\begin{cases}
0, & \text{if } \text{CarPosition}_t \geq \text{HillDistances}[1] + \text{HillHeights}[1] \\
-1, & \text{otherwise}
\end{cases}
\end{equation}
where $R_t$: Reward at time $t$; $\text{CarPosition}_t$: The car's position at time $t$; $\text{HillDistances}[1] + \text{HillHeights}[1]$: The position of the top of the right hill.
\end{tcolorbox}

Based on the extracted JSON specification, we trained the generated RL code, and the resulting training loss curve and policy evaluation outcomes are shown below.

\begin{figure}[ht]
    \begin{center}
    \subfigure[Training progress of DQN by A-LAMP]{
        \includegraphics[width=0.45\textwidth]{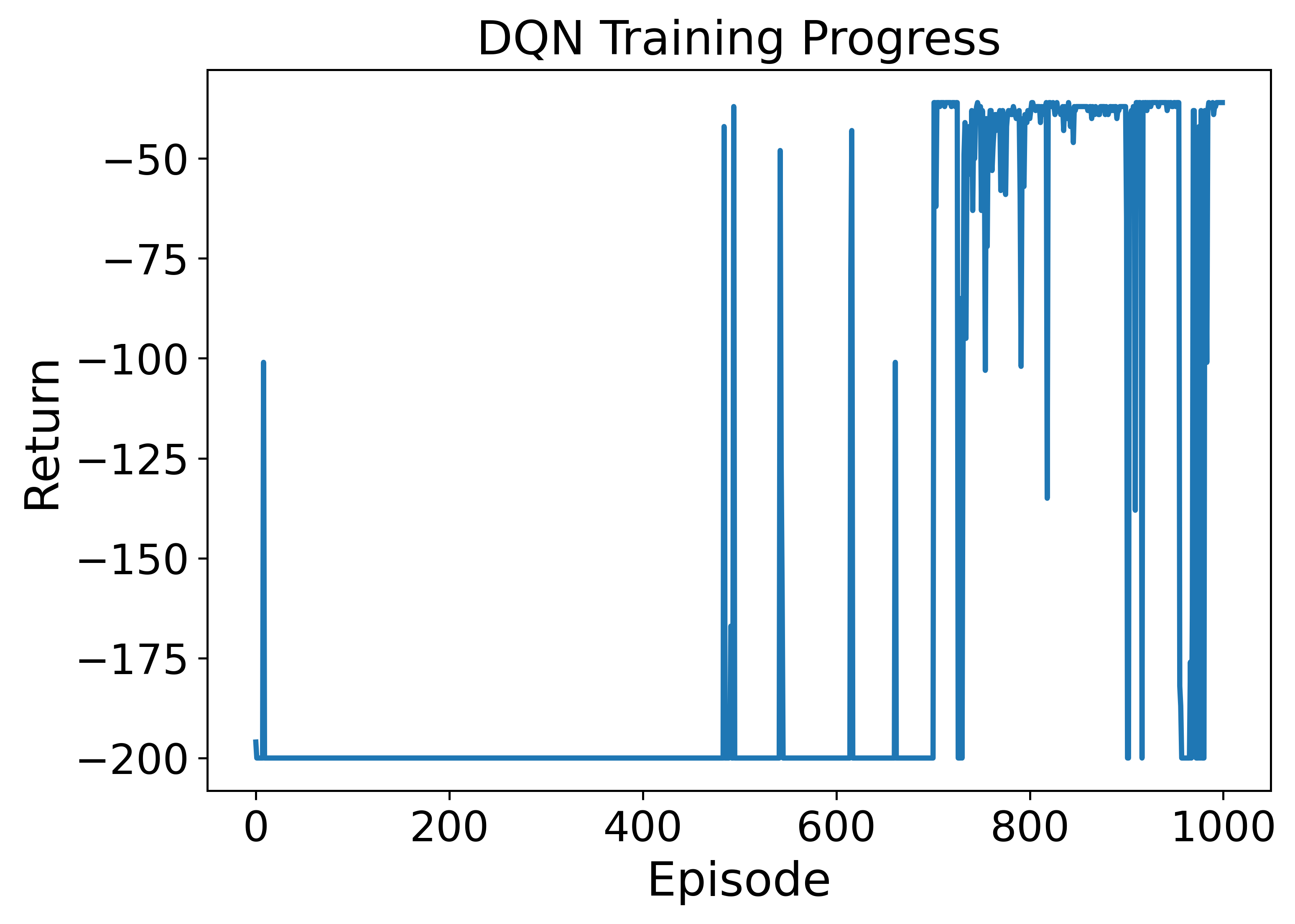}
    }
    \hfil
    \subfigure[Policy evaluation: DQN vs. Optimal baseline.]{
        \centering
        \includegraphics[width=0.45\textwidth]{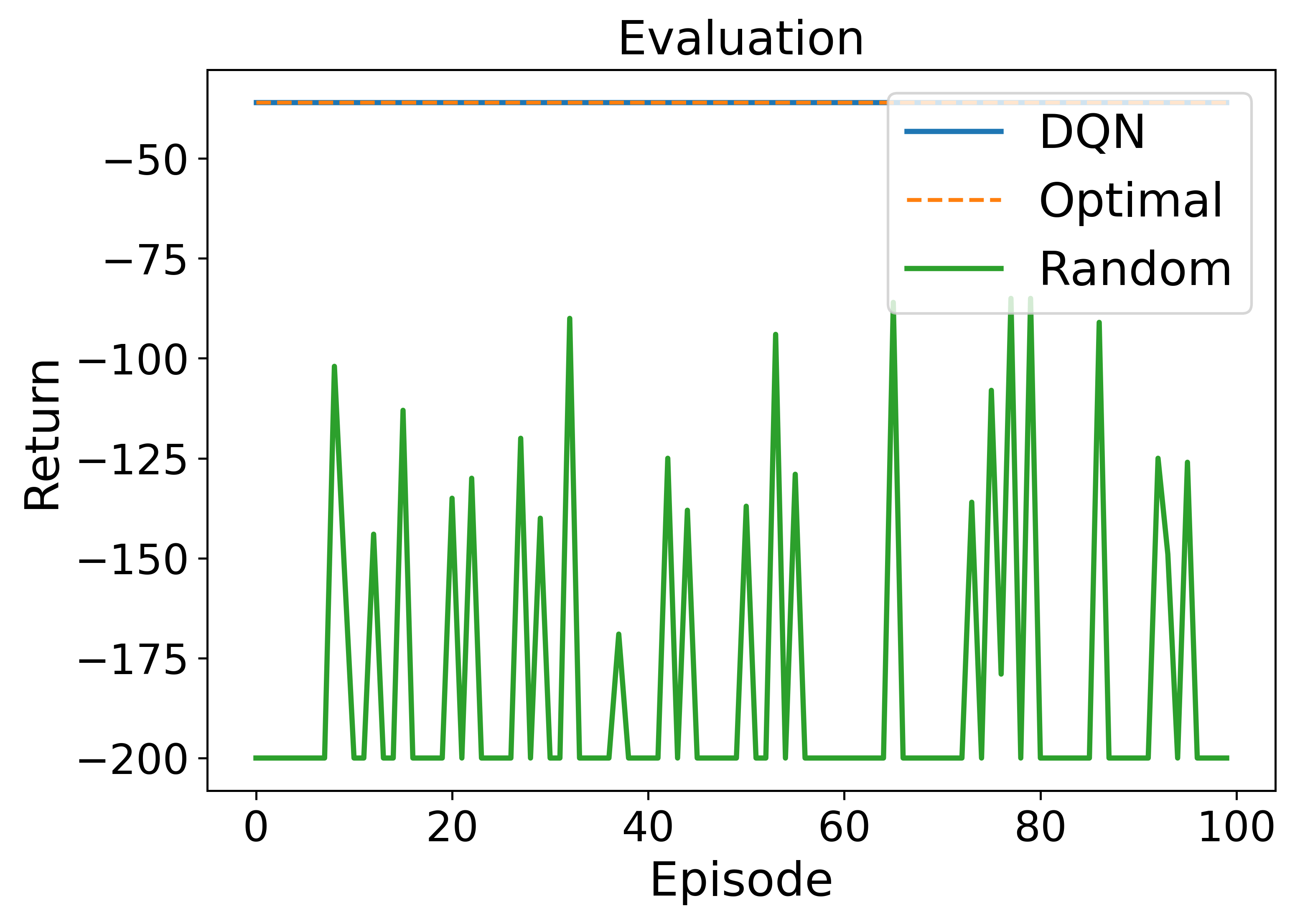}
    }
    \caption{Case study. A-LAMP successfully generates an executable environment enabling stable DQN training and competitive policy performance.}
    \end{center}
\end{figure}

\subsection{Case Study: Wireless}

\begin{tcolorbox}[title=Parameters (JSON)]
\begin{itemize}
  \item \textbf{Bandwidth}: The bandwidth of the system in MHz (float).
  \item \textbf{TransmissionPower}: The transmission power in mW (float).
  \item \textbf{NoiseDensity}: The noise density in dBm (float).
  \item \textbf{PathLossCoefficient}: Path loss coefficient affecting the channel (float).
  \item \textbf{ShadowingStandardDeviation}: Standard deviation of log-normal shadowing in dB (float).
  \item \textbf{UserDistances}: Distances of users from the base station (array of size 4, in meters).
  \item \textbf{ChannelGainRange}: Normalized range of channel gains in dB (array of size 2).
\end{itemize}
\end{tcolorbox}

\begin{tcolorbox}[title=Variables (JSON)]
\begin{itemize}
  \item \textbf{ScheduledUser}: Index of the user scheduled to transmit in a time slot.
  \item \textbf{TransmissionRate}: Transmission rate for the scheduled user in each time slot.
  \item \textbf{ChannelGain}: Time-varying channel gain for each user.
\end{itemize}
\end{tcolorbox}

\begin{tcolorbox}[title=Objective (JSON)]
The goal is to determine an optimal scheduling strategy that maximizes overall network performance while adhering to system constraints.  
Formally:
\begin{equation}
  J(\pi) = \mathbb{E} \left[ \sum_{t=0}^{\infty} \gamma^t \cdot 
  \left( \text{B} \cdot \log_2 \left( 1 + \frac{\text{TransmissionPower} \cdot \text{ChannelGain}[t]}{\text{NoiseDensity} \cdot \text{Bandwidth}} \right) \right) \;\middle|\; \pi \right]
\end{equation}
\end{tcolorbox}

\begin{tcolorbox}[title=Constraints (JSON)]
\begin{itemize}
  \item Only one user can be scheduled per time slot.
  \begin{equation}
    \forall t, \quad \sum_{u} x_{u,t} = 1
  \end{equation}

  \item Transmission rate must follow Shannon’s formula.
  % \begin{equation}
  %   \forall t, \quad 
  %   \text{TransmissionRate}[t] = \text{Bandwidth} \cdot \log_2 \left( 1 + \frac{\text{TransmissionPower} \cdot \text{ChannelGain}[t]}{\text{NoiseDensity} \cdot \text{Bandwidth}} \right)
  % \end{equation}

\begin{equation}
\begin{aligned}
\forall t,\quad 
\text{TransmissionRate}[t] 
&= \text{Bandwidth} \cdot \\
&\quad \log_2 \Biggl(
        1 +
        \frac{
            \text{TransmissionPower} \cdot \text{ChannelGain}[t]
        }{
            \text{NoiseDensity} \cdot \text{Bandwidth}
        }
    \Biggr)
\end{aligned}
\end{equation}

  \item Channel gain values must be between $-80$ dB and $-30$ dB.
  \begin{equation}
    -80 \leq \text{ChannelGain}[t] \leq -30
  \end{equation}

  \item Transmission power must be non-negative.
  \begin{equation}
    \text{TransmissionPower} \geq 0
  \end{equation}

  \item Bandwidth must be non-negative.
  \begin{equation}
    \text{Bandwidth} \geq 0
  \end{equation}

  \item Noise density must be non-negative.
  \begin{equation}
    \text{NoiseDensity} \geq 0
  \end{equation}

  \item User distances must be non-negative.
  \begin{equation}
    \text{UserDistances}[i] \geq 0
  \end{equation}
\end{itemize}
\end{tcolorbox}

\begin{tcolorbox}[title=Environment (JSON)]
This task requires a custom Gym environment.  
Transition logic: Schedule a user, compute transmission rate using Shannon’s formula based on current channel gain, update the environment state, and proceed to the next time step.  
Channel gain is assumed to be updated externally each time step.
\end{tcolorbox}

\begin{tcolorbox}[title=State (JSON)]
The state space includes the channel gain, path loss coefficient, shadowing, and user distances from the base station.  
Variables: \textbf{ChannelGain}, \textbf{PathLossCoefficient}, \textbf{ShadowingStandardDeviation}, \textbf{UserDistances}.  
Shape: [4,].
\end{tcolorbox}

\begin{tcolorbox}[title=Action (JSON)]
The action space is discrete and consists of selecting one user among the available four to schedule.  
Variables: \textbf{ScheduledUser}.  
Shape: [4].  
Type: discrete.
\end{tcolorbox}

\begin{tcolorbox}[title=Reward (JSON)]
The reward is the transmission rate achieved using Shannon’s capacity formula:  
\begin{equation}
  R_t = \text{B} \cdot \log_2 \left( 1 + \frac{\text{TransmissionPower} \cdot \text{ChannelGain}[t]}{\text{NoiseDensity} \cdot \text{Bandwidth}} \right)
\end{equation}
where $R_t$: reward at time $t$; $\text{B}$: system bandwidth; $\text{TransmissionPower}$: transmission power in mW; $\text{ChannelGain}[t]$: channel gain for the scheduled user; $\text{NoiseDensity}$: noise density in dBm; $\text{Bandwidth}$: transmission bandwidth in MHz.
\end{tcolorbox}

\subsection{Case Study: Drone Delivery}

\begin{tcolorbox}[title=Parameters (JSON)]
\begin{itemize}
  \item \textbf{GridSize}: $50 \times 50$ grid world
  \item \textbf{InitialEnergyMin}: Minimum initial energy level
  \item \textbf{InitialEnergyMax}: Maximum initial energy level
  \item \textbf{MovementEnergyCost}: Energy cost for moving to an adjacent cell
  \item \textbf{DeliveryEnergyCost}: Energy cost for each package delivery
  \item \textbf{NumberOfPackages}: Number of packages to deliver, sampled per episode
  \item \textbf{PickupLocations}: Coordinates of pickup points
  \item \textbf{DeliveryTargets}: Coordinates of delivery destinations
\end{itemize}
\end{tcolorbox}

\begin{tcolorbox}[title=Variables (JSON)]
\begin{itemize}
  \item \textbf{Route}: Sequence of coordinates the drone visits [S,2]
  \item \textbf{PickupOrder}: Order in which packages are picked up
  \item \textbf{DeliveryOrder}: Order in which packages are delivered
  \item \textbf{EnergyConsumed}: Total energy consumed by moves and deliveries
  \item \textbf{CurrentEnergy}: Energy level at a given state
  \item \textbf{CurrentPosition}: Drone’s current position in the grid
\end{itemize}
\end{tcolorbox}

\begin{tcolorbox}[title=Objective (JSON)]
The objective is to complete all deliveries while minimizing energy consumption, ensuring that energy is not depleted prematurely.  
Formally:
\begin{equation}
  J(\pi) = \mathbb{E}_{\pi} \left[ \sum_{t=0}^{T} \gamma^t R_t \right]
\end{equation}
where $R_t$ denotes the reward at time $t$, and $\gamma$ is the discount factor.
\end{tcolorbox}

\begin{tcolorbox}[title=Constraints (JSON)]
\begin{itemize}
  \item Energy consumed must not exceed the maximum initial energy.
  \begin{equation}
    \text{EnergyConsumed} \leq \text{InitialEnergyMax}
  \end{equation}

  \item Route must include all pickup locations.
  \begin{equation}
    \forall p, \exists s \; \text{such that } \text{Route}[s] = \text{PickupLocations}[p]
  \end{equation}

  \item Route must include all delivery targets.
  \begin{equation}
    \forall d, \exists t \; \text{such that } \text{Route}[t] = \text{DeliveryTargets}[d]
  \end{equation}

  \item Route must start at the initial position.
  \begin{equation}
    \text{Route}[0] = \text{CurrentPosition}
  \end{equation}

  \item Energy consumed must equal the number of moves between adjacent cells.
  \begin{equation}
    \text{EnergyConsumed} = \sum_{i=1}^{S-1} \mathbb{I}\big(\|\text{Route}[i] - \text{Route}[i+1]\|_1 = 1\big)
  \end{equation}

  \item Current energy must remain non-negative throughout the route.
  \begin{equation}
    \text{CurrentEnergy}[s] \geq 0
  \end{equation}

  \item Each package must be picked up before being delivered.
  \begin{equation}
    \text{PickupOrder}[i] < \text{DeliveryOrder}[i]
  \end{equation}

  \item The number of packages picked up and delivered must match the total number of packages.
  \begin{equation}
    \text{NumberOfPackagesPicked} = \text{NumberOfPackages}
  \end{equation}
  \begin{equation}
    \text{NumberOfPackagesDelivered} = \text{NumberOfPackages}
  \end{equation}

  \item All pickup and delivery coordinates must lie within the $50 \times 50$ grid.
  \begin{equation}
    0 \leq \text{PickupLocations}[i,j] < 50
  \end{equation}
  \begin{equation}
    0 \leq \text{DeliveryTargets}[i,j] < 50
  \end{equation}
\end{itemize}
\end{tcolorbox}

\begin{tcolorbox}[title=Environment (JSON)]
This task requires a custom Gym environment.  
Transition logic: When an action is taken, update the drone's position, decrease energy by movement or delivery cost, and adjust the reward if a delivery is completed.
\end{tcolorbox}

\begin{tcolorbox}[title=State (JSON)]
The state includes current position, energy, and delivery/pickup status.  
Variables: \textbf{CurrentPosition}, \textbf{CurrentEnergy}, \textbf{PickupStatuses}, \textbf{DeliveryStatuses}.  
Shape: [1 + 2 + n + n].
\end{tcolorbox}

\begin{tcolorbox}[title=Action (JSON)]
The action space consists of 6 discrete actions: \textbf{MoveNorth}, \textbf{MoveSouth}, \textbf{MoveEast}, \textbf{MoveWest}, \textbf{PickUpPackage}, \textbf{DeliverPackage}.  
Shape: [6].  
Type: discrete.
\end{tcolorbox}

\begin{tcolorbox}[title=Reward (JSON)]
The reward encourages efficient deliveries while penalizing energy use.  
Formally:
\begin{equation}
R_t =
\begin{cases}
+10, & \text{if a package is delivered} \\
-1 \times \text{MovementEnergyCost}, & \text{for each move} \\
-2 \times \text{DeliveryEnergyCost}, & \text{for each delivery} \\
0, & \text{otherwise}
\end{cases}
\end{equation}
\end{tcolorbox}

Based on the extracted JSON specification, we trained the generated RL code, and the resulting training loss curve and policy evaluation outcomes are shown below.

\begin{figure}[ht]
    \begin{center}
    \subfigure[Training progress of DQN by A-LAMP]{
        \includegraphics[width=0.45\textwidth]{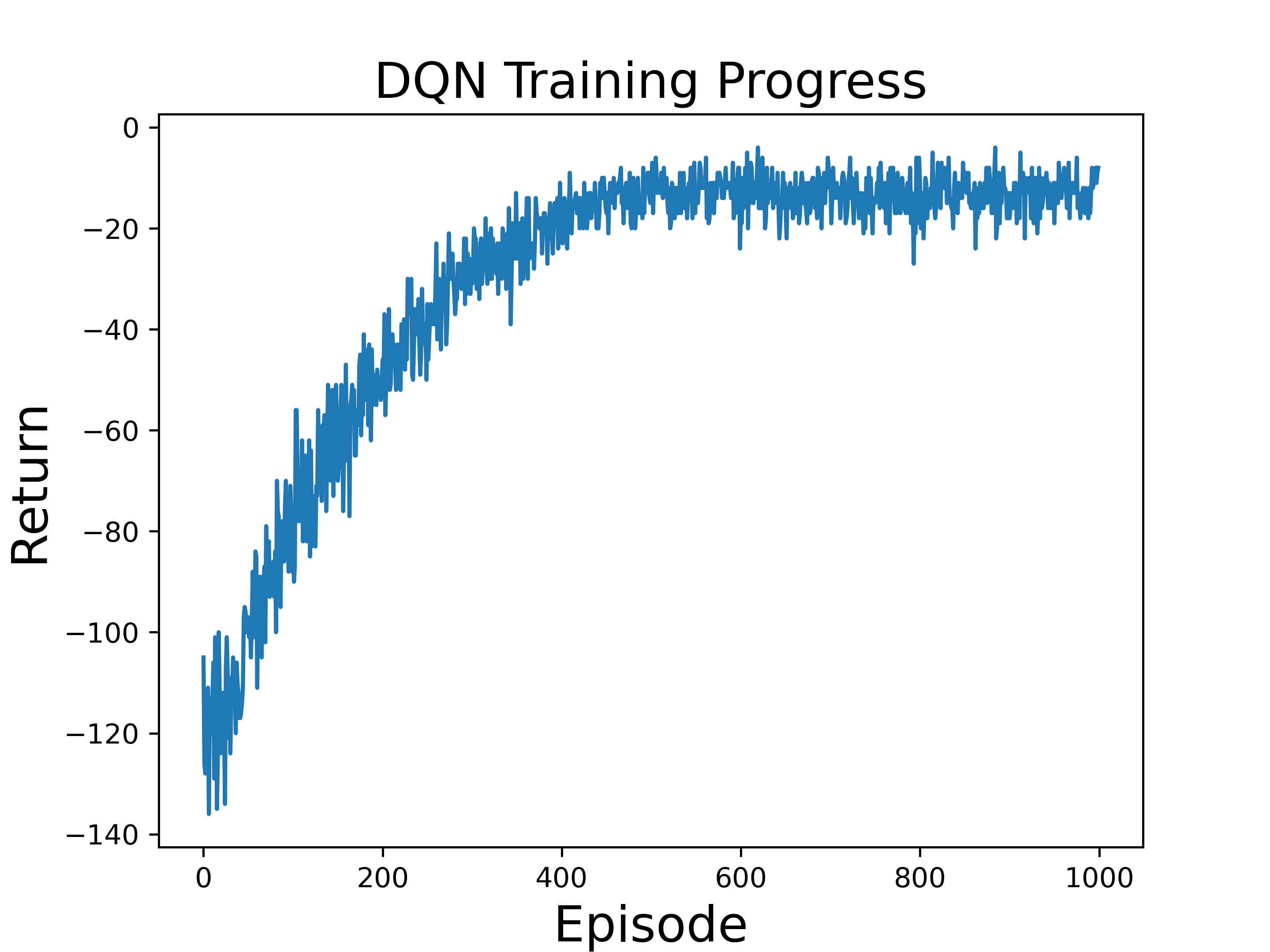}
    }
    \hfil
    \subfigure[Policy evaluation: DQN vs. Optimal baseline.]{
        \centering
        \includegraphics[width=0.45\textwidth]{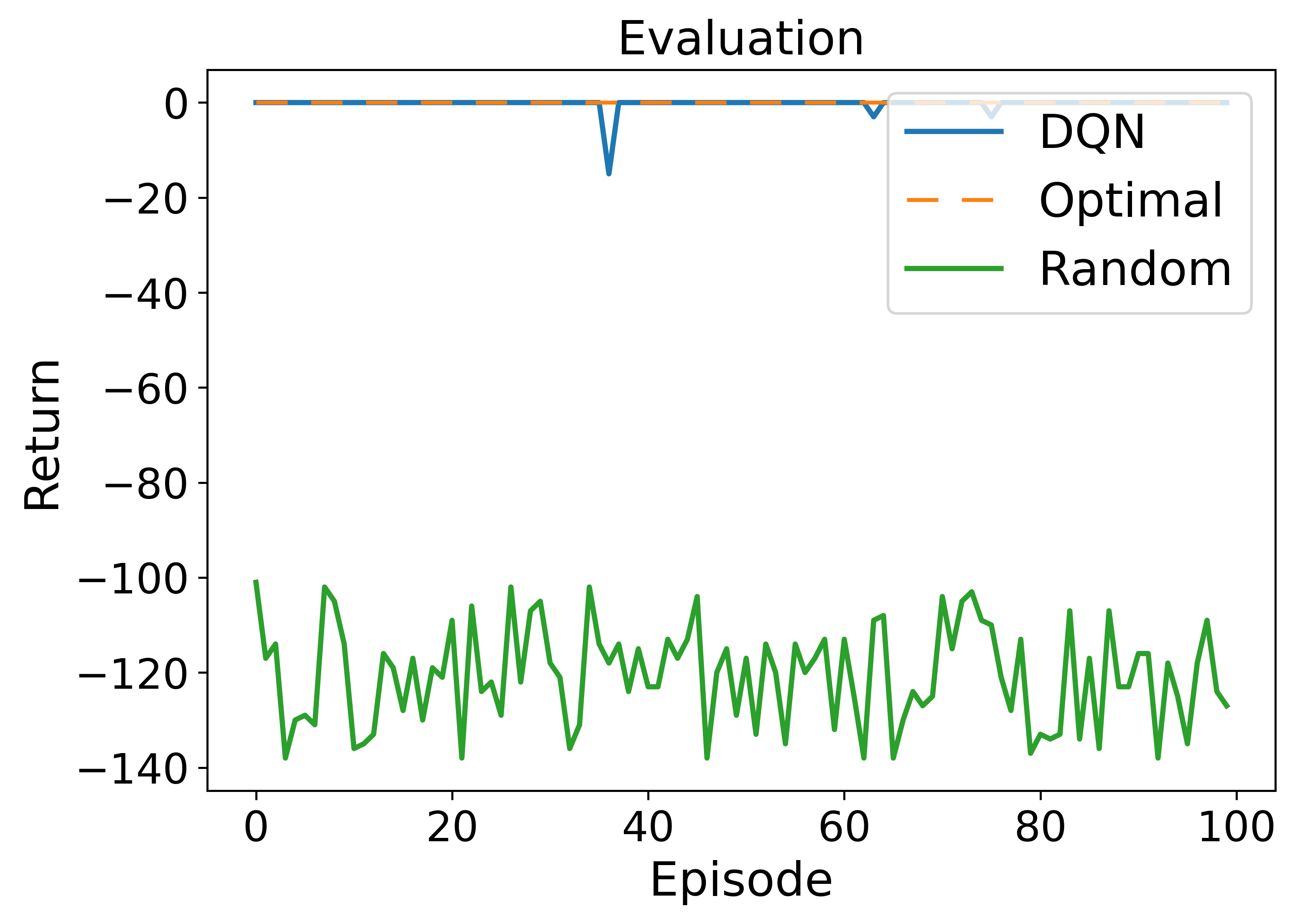}
    }
    \caption{Case study. A-LAMP successfully generates an executable environment enabling stable DQN training and competitive policy performance.}
    \end{center}
\end{figure}

\subsection{Case Study: Inventory-Management}

\begin{tcolorbox}[title=Parameters (JSON)]
\begin{itemize}
  \item \textbf{NumberOfItems}: The number of items in the inventory (int).
  \item \textbf{FixedCost}: Fixed ordering cost for each item (float, shape [10]).
  \item \textbf{UnitCost}: Per-unit ordering cost for each item (float, shape [10]).
  \item \textbf{HoldingCost}: Cost of holding inventory per time unit for each item (float, shape [10]).
  \item \textbf{SellingPrice}: Selling price per unit for each item (float, shape [10]).
  \item \textbf{MeanDemand}: Mean of the Poisson distribution for demand for each item (int, shape [10]).
\end{itemize}
\end{tcolorbox}

\begin{tcolorbox}[title=Variables (JSON)]
\begin{itemize}
  \item \textbf{OrderQuantity}: Quantity of each item to order at each time step (shape [10]).
  \item \textbf{StockLevel}: Current inventory level for each item (shape [10]).
  \item \textbf{LostSales}: Unmet demand due to insufficient stock (shape [10]).
  \item \textbf{Profit}: Long-term profit to be maximized (scalar).
\end{itemize}
\end{tcolorbox}

\begin{tcolorbox}[title=Objective (JSON)]
The goal is to maximize long-term profit while fulfilling demand.  
Formally:
% \begin{equation}
%   J(\pi) = \mathbb{E} \left[ \sum_{t=0}^{\infty} \gamma^t \cdot 
%   \left( \sum_{i=1}^{10} \left( 
%   \text{SellingPrice}_i \cdot \min(\text{Demand}_i(t), \text{StockLevel}_i(t))
%   - \text{FixedCost}_i
%   - \text{UnitCost}_i \cdot \text{OrderQuantity}_i(t)
%   - \text{HoldingCost}_i \cdot \text{StockLevel}_i(t)
%   \right) \right) \;\middle|\; \pi \right]
% \end{equation}
\begin{equation}
\begin{aligned}
J(\pi) = \mathbb{E} \Biggl[ \sum_{t=0}^{\infty} \gamma^t \cdot \Biggl( 
    \sum_{i=1}^{10} \Big( 
    &\text{SellingPrice}_i \cdot \min(\text{Demand}_i(t), \text{StockLevel}_i(t)) \\
    &- \text{FixedCost}_i 
    - \text{UnitCost}_i \cdot \text{OrderQuantity}_i(t) \\
    &- \text{HoldingCost}_i \cdot \text{StockLevel}_i(t)
    \Big) \Biggr) \;\Big|\, \pi \Biggr]
\end{aligned}
\end{equation}

\end{tcolorbox}

\begin{tcolorbox}[title=Constraints (JSON)]
\begin{itemize}
  \item Order quantity for each item must be non-negative.
  \begin{equation}
    \text{OrderQuantity}[i] \geq 0
  \end{equation}

  \item Inventory levels must be non-negative.
  \begin{equation}
    \text{StockLevel}[i] \geq 0
  \end{equation}

  \item Demand follows a Poisson distribution with mean $8$.
  \begin{equation}
    \text{Demand}[i] \sim \text{Poisson}(\lambda=8)
  \end{equation}

  \item Lost sales penalties apply when demand exceeds stock.
  \begin{equation}
    \text{LostSales}[i] = \max(\text{Demand}[i] - \text{StockLevel}[i], 0)
  \end{equation}
\end{itemize}
\end{tcolorbox}

\begin{tcolorbox}[title=Environment (JSON)]
This environment requires a custom Gym implementation.  
Transition logic: For each item, update stock levels by fulfilling demand and applying order quantities.  
Calculate lost sales when demand exceeds stock, then compute costs and update profit.
\end{tcolorbox}

\begin{tcolorbox}[title=State (JSON)]
The state space includes current stock levels and demand for each item.  
Variables: \textbf{StockLevel}, \textbf{Demand}.  
Shape: [20,].
\end{tcolorbox}

\begin{tcolorbox}[title=Action (JSON)]
The action space is discrete and consists of order quantities for each of the $10$ items.  
Variables: \textbf{OrderQuantity}.  
Shape: [10].  
Type: discrete.
\end{tcolorbox}

\begin{tcolorbox}[title=Reward (JSON)]
The reward is the profit gained at each time step, defined as:
% \begin{equation}
% R_t = \sum_{i=1}^{10} \Big( 
%   \text{SellingPrice}_i \cdot \min(\text{Demand}_i(t), \text{StockLevel}_i(t)) 
%   - \text{FixedCost}_i \cdot \mathbb{I}(\text{OrderQuantity}_i(t) > 0)
%   - \text{UnitCost}_i \cdot \text{OrderQuantity}_i(t)
%   - \text{HoldingCost}_i \cdot \text{StockLevel}_i(t)
%   - \text{PenaltyCost}_i \cdot \max(\text{Demand}_i(t) - \text{StockLevel}_i(t), 0)
% \Big)
% \end{equation}

\begin{equation}
\begin{aligned}
R_t = \sum_{i=1}^{10} \Big(
    &\text{SellingPrice}_i \cdot \min(\text{Demand}_i(t), \text{StockLevel}_i(t))
    - \text{FixedCost}_i \cdot  \\
    &  \mathbb{I}(\text{OrderQuantity}_i(t) > 0)- \text{UnitCost}_i \cdot \text{OrderQuantity}_i(t)
    - \text{HoldingCost}_i \cdot  \\
    & \text{StockLevel}_i(t) - \text{PenaltyCost}_i \cdot \max(\text{Demand}_i(t) - \text{StockLevel}_i(t), 0)
\Big)
\end{aligned}
\end{equation}

where $R_t$: profit at time $t$; $\mathbb{I}(\cdot)$: indicator function for ordering; $\text{PenaltyCost}$: penalty for unmet demand.
\end{tcolorbox}

Based on the extracted JSON specification, we trained the generated RL code, and the resulting training loss curve and policy evaluation outcomes are shown below.

\begin{figure}[ht]
    \begin{center}
    \subfigure[Training progress of DQN by A-LAMP]{
        \includegraphics[width=0.45\textwidth]{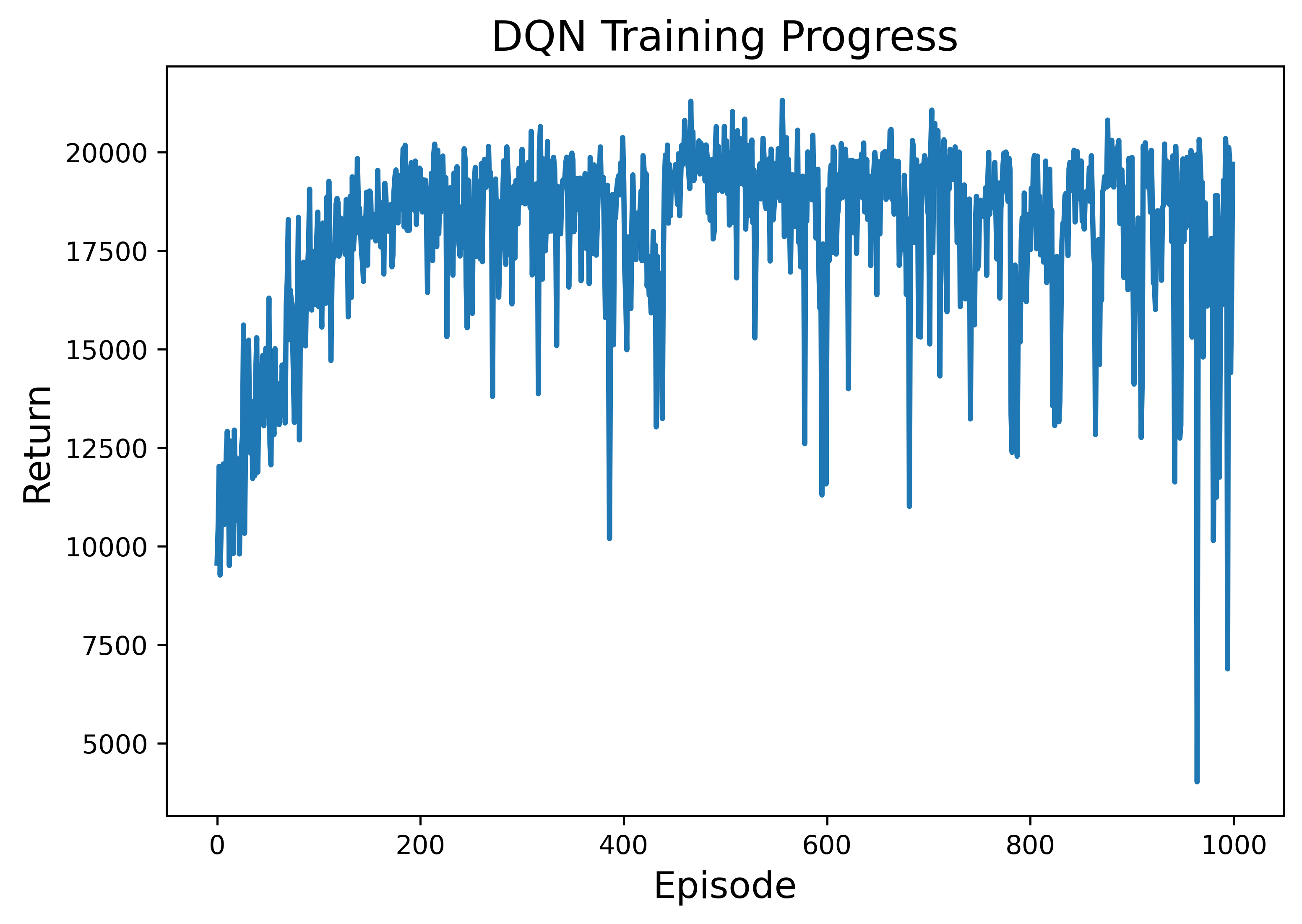}
    }
    \hfil
    \subfigure[Policy evaluation: DQN vs. Optimal baseline.]{
        \centering
        \includegraphics[width=0.45\textwidth]{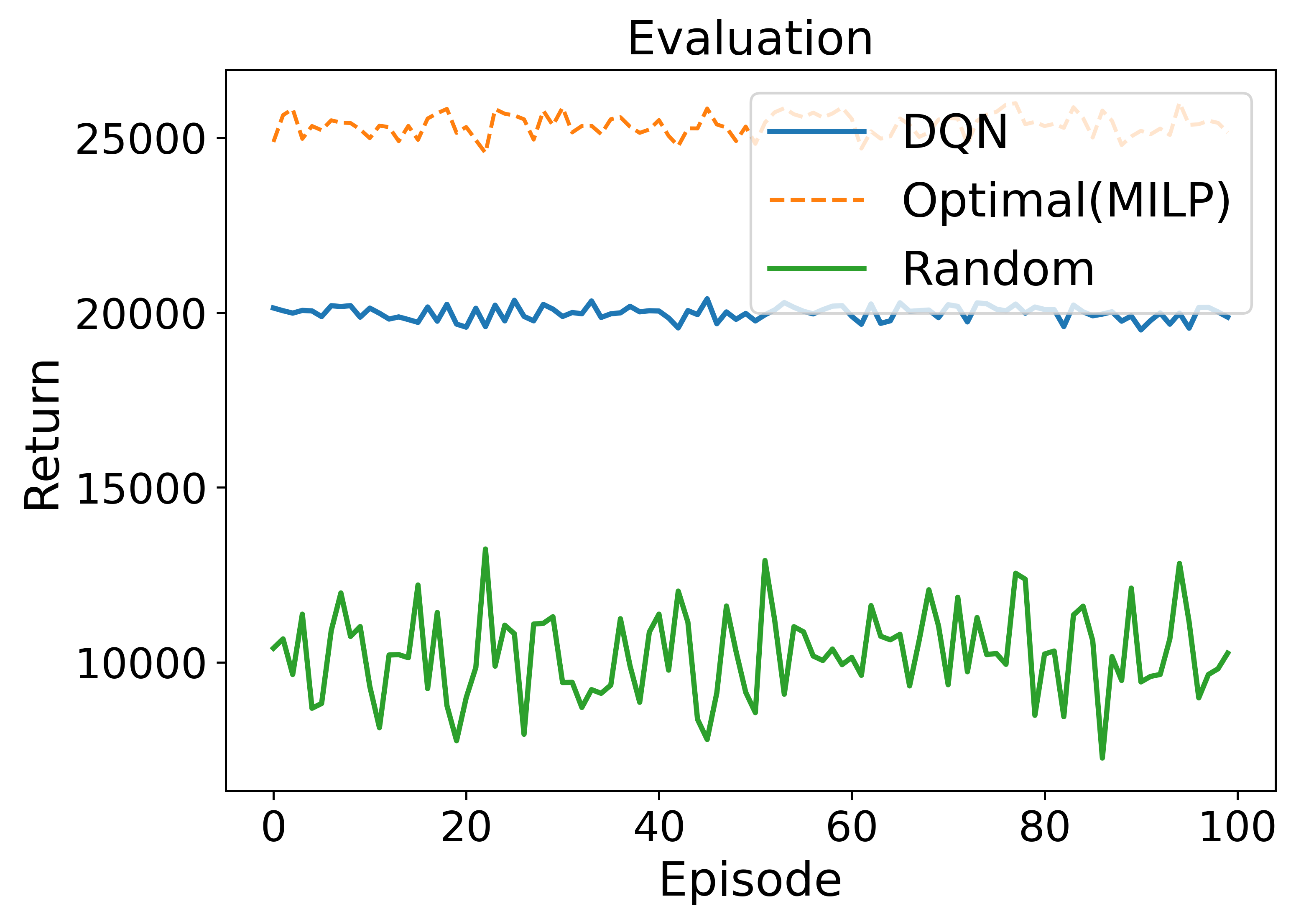}
    }
    \caption{Case study. A-LAMP successfully generates an executable environment enabling stable DQN training and competitive policy performance.}
    \end{center}
\end{figure}

\section{Prompts for Agents}
\label{sec:prompts}

Below are partial examples of the prompts used for each agent.

\begin{tcolorbox}[title=Parameter Agent Prompt, breakable]
Here is the natural language description of an optimization problem: \\

----- \\
\{description\} \\
----- \\

Your task is to identify and extract parameters from the description. \\
The parameters are values that are already known. \\
Please generate the output in the following format: \\

\[
\vdots
\]

Where SYMBOL is a string representing the parameter (use CamelCase), \\
SHAPE is the shape of the parameter (e.g. "[]" for scalar, or "[N, M]" for a matrix of size \(N \times M\) \\
where \(N\) and \(M\) are scalar parameters), \\
DEFINITION is a string describing the parameter, and TYPE is one of "int", "float", or "binary". \\

\[
\vdots
\]

- Put all the parameters in a single json object. \\
- Do not generate anything after and before the json object. \\
Take a deep breath and think step by step. \\
\end{tcolorbox}

\begin{tcolorbox}[title=Objective Agent Prompt]
Here is the natural language description of an optimization problem: \\

----- \\
\{description\} \\
----- \\

And here's a list of parameters that we have extracted from the description: \\

\{params\} \\

Your task is to identify and extract the optimization objective from the description. \\
The objective is the goal that the optimization model is trying to achieve (e.g. maximize profit, minimize cost). \\
The objective will be used in MDP. \\
Please generate the output in the following format: \\

===== \\
OBJECTIVE: objective description \\
===== \\

\[
\vdots
\]

- Do not generate anything after and before the objective. \\
Take a deep breath and think step by step. \\
\end{tcolorbox}

\begin{tcolorbox}[title=Variable Agent Prompt]
Here is the natural language description of an optimization problem: \\

----- \\
\{description\} \\
----- \\

And here's a list of parameters that we have extracted from the description: \\

----- \\
\{params\} \\
----- \\

Your task is to identify and extract variables from the description. \\
The variables are values that are not known and need to be determined by the optimization model. \\
Please generate the output in the following format: \\

\[
\vdots
\]

Where SYMBOL is a string representing the variable (use CamelCase), \\
SHAPE is the shape of the variable (e.g. "[]" for scalar, or "[N, M]" for a matrix of size \(N \times M\)), \\
and DEFINITION is a string describing the variable. \\

\[
\vdots
\]

- Put all the parameters in a single json object. \\
- Do not generate anything after and before the json object. \\
Take a deep breath and think step by step. \\
\end{tcolorbox}

\begin{tcolorbox}[title=Objective Modeling Agent Prompt, breakable]
Here is the natural language description of an optimization problem: \\

----- \\
\{description\} \\
----- \\

Parameters: \{params\} \\
Variables: \{vars\} \\
Constraints: \{constraints\} \\

Your task is to model the following objective mathematically in LaTeX for the MDP formulation: \\

\{objective\} \\

MDP objective formula will be Expectation of action that is chosen at each time step. \\
Please generate the output in the following format: \\

===== \\
objective formulation in LaTeX, between \$...\$, \\
===== \\

\[
\vdots
\]

- You can only use existing parameters and variables in the formulation. \\
- But you can change the shape of variable and parameters. \\
- Do not generate anything after and before the objective. \\
Take a deep breath and think step by step. \\
\end{tcolorbox}

\begin{tcolorbox}[title=Constraints Modeling Agent Prompt]
You are an expert in optimization modeling. \\

Here is the natural language description of an optimization problem: \\

----- \\
\{description\} \\
----- \\

Parameters: \{params\} \\
Variables: \{vars\} \\

Your task is to model the following constraint mathematically in LaTeX for the MDP formulation: \\

\{constraints\} \\

The constraints are the conditions that must be satisfied by the variables. \\
Please generate the output in the following format: \\

===== \\
constraint formulation in LaTeX, between \$...\$, \\
===== \\

\[
\vdots
\]

- You can only use existing parameters and variables in the formulation. \\
- Do not generate anything after and before the constraint. \\
Take a deep breath and think step by step. \\
\end{tcolorbox}

\begin{tcolorbox}[title=SAR Agent Prompt]
You are an expert in reinforcement learning and scheduling optimization. \\
Your task is to extract key components for designing a Deep Q-Network (DQN) scheduler. \\

Here is the natural language description of the scheduling problem: \\
----- \\
\{description\} \\
----- \\

Parameters: \{params\} \\
Variables: \{vars\} \\
Constraints: \{constraints\} \\
Objective: \{objective\} \\

Your task is to identify and define the following components for reinforcement learning: \\

1. State Space \\
The state space represents the environment's status at a single time step. \\

\[
\vdots
\]

2. Action Space \\
The action space is defined as the set of all possible actions that the agent can take. \\

\[
\vdots
\]

3. Reward Function \\
The reward function quantifies the quality of the agent's decision. \\

\[
\vdots
\]

Take a deep breath and think step by step. \\
\end{tcolorbox}

\begin{tcolorbox}[title=Env Agent Prompt]
You are an expert in reinforcement learning and scheduling optimization. \\
Your task is to define the transition dynamics for a DQN-based scheduler environment without using OpenAI Gym unless a matching Gym environment already exists. \\

You will be given the following: \\
- Natural language description: \{description\} \\
- Parameters: \{params\} \\
- Variables: \{vars\} \\
- Reward: \{reward\} \\
- Constraints: \{constraints\} \\
- Objective: \{objective\} \\

1. **Check for Existing Gym Environment** \\
2. **Extract Transition Dynamics** \\
3. **Output Format** (JSON) \\

\[
\vdots
\]

- Do not redefine states, actions, or rewards. \\
- Keep the JSON output clean. \\
Take a deep breath and think step by step. \\
\end{tcolorbox}

\begin{tcolorbox}[title=Coding Agent Prompt]
You are a PyTorch-based Deep Reinforcement Learning expert. \\
Your task is to generate a complete Deep Q-Network (DQN) training implementation using PyTorch, based strictly on the structured MDP information. \\

Environment: \{env\} \\
Parameters: \{params\} \\
State/Action: \{vars\} \\
Reward: \{reward\} \\
Constraints: \{constraints\} \\
Objective: \{objective\} \\

Your tasks: \\
1. Implement a \texttt{CustomEnv}. \\
2. Implement Q-Network. \\
3. Training loop (1000 episodes, replay buffer, target updates). \\
4. Save model and training results. \\

\[
\vdots
\]

- Only use PyTorch and standard libraries. \\
- Ensure consistent tensor handling. \\
- Do not generate anything before and after the code. \\
Take a deep breath and think step by step. \\
\end{tcolorbox}

\end{document}